%% file: CADForge.tex
\documentclass{article} 
\usepackage{iclr2027_conference,times}

\input{math_commands.tex}

\usepackage{hyperref}
\usepackage{url}
\usepackage{booktabs}
\usepackage{graphicx}
\usepackage{wrapfig}
\usepackage[most]{tcolorbox}
\usepackage{enumitem}

\title{CADForge: Agentic Single-View CAD Reconstruction with Explicit Geometry Reasoning}

\author{
Keyang Lu$^1$ \enspace
Zhifei Yang$^1$ \enspace
Tianao Dong$^2$ \enspace
Mingzhe Xing$^1$ \enspace
Zhen Xiao$^{1*}$ \enspace
Yikai Wang$^{3*}$ \\
\normalfont
$^1$Peking University \quad
$^2$Shenzhen Loop Area Institute \quad
$^3$Beijing Normal University \\
\normalfont
$^*$ Corresponding Authors
}

\iclrfinalcopy 
\begin{document}

\maketitle
\lhead{CADForge: Agentic Single-View CAD Reconstruction with Explicit Geometry Reasoning}

\input{Sections/0_abstract}

\input{Sections/1_introduction}

\input{Sections/2_related_work}

\input{Sections/3_method}

\input{Sections/4_experiments}

\input{Sections/5_conclusion}

\bibliography{iclr2027_conference}
\bibliographystyle{iclr2027_conference}

\appendix
\input{Sections/X_supplementary}

\end{document}

%% file: math_commands.tex
\usepackage{amsmath,amsfonts,bm}

\def\eqref#1{equation~\ref{#1}}

\def\1{\bm{1}}

\DeclareMathAlphabet{\mathsfit}{\encodingdefault}{\sfdefault}{m}{sl}
\SetMathAlphabet{\mathsfit}{bold}{\encodingdefault}{\sfdefault}{bx}{n}



%% file: Sections/0_abstract.tex
\begin{abstract}
Reconstructing editable parametric CAD models from a single-view image is of great practical value for modern manufacturing, yet remains challenging due to incomplete geometric observations and complex inter-part relationships. To address it, we propose \textit{\textbf{CADForge}}, an agentic framework that progressively converts a single image into CadQuery programs. 
CADForge decomposes an object into CAD-meaningful components and performs explicit geometric reasoning for each component, a process that first identifies CAD-relevant constraints and then translates them into precise modeling parameters through mathematical code.
The inferred parameters then drive component-wise synthesis of executable CadQuery programs, with a review agent evaluating the resulting geometry and providing targeted feedback for iterative refinement. To further improve robustness and efficiency, CADForge incorporates a failure-guided toolkit construction mechanism to distill accumulated experience into tools, and maintains a compact parametric CAD memory for retrieving modeling context on demand. Experiments on diverse single- and multi-part objects show that CADForge consistently outperforms existing baselines in reconstruction fidelity and perceptual quality, demonstrating an effective approach to accurate single-view CAD reconstruction.

\end{abstract}

%% file: Sections/1_introduction.tex
\section{Introduction}
Computer-aided design (CAD) plays an essential role in modern engineering (\cite{narayan2013computer}), where 3D objects can be constructed through sequences of modeling operations and editable parameters. Compared with point clouds and meshes, parametric CAD models can provide precise boundary representations (B-Reps) and flexible editability, making them well suited for real-world applications such as furniture design (\cite{scurtu2015optimization}), and industrial manufacturing (\cite{campi2022cad}). However, creating such models typically requires specialized expertise and considerable manual effort, driving growing interest in automatic parametric CAD generation.

Earlier work explored automatic CAD generation by learning neural representations of CAD construction sequences with dedicated network architectures (\cite{wu2021deepcad,xu2022skexgen,xu2023hierarchical}). With advances in large language models (LLMs) and vision-language models (VLMs), current work has made considerable progress in generating parametric CAD models from text descriptions (\cite{khan2024text2cad,li2025cad,guan2026cad}), point clouds (\cite{rukhovich2025cad,kabisov2026cadreasoner, li2026pllm}) or multi-view images (\cite{kolodiazhnyi2025cadrille,Pyatov_2026_CVPR}), achieving strong performance in both geometric accuracy and structural validity. 

However, accurate parametric CAD reconstruction from a single-view image, which offers significant practical value in industrial design and 3D printing (\cite{sachs1990three}), remains challenging. This task is inherently nontrivial as it requires a model not only to infer complete 3D geometry and structural relationships from partial visual observations but also to translate them into precise CAD modeling parameters and commands. Existing methods (\cite{li2025caddreamer,noeckel2025view2cad,chen2025cadcrafter,doris2026cad}) primarily rely on task-specific fine-tuning over curated image-CAD datasets and struggle to produce complex multi-part objects like furniture. When applied to in-the-wild images, they often produce distorted geometry, omit important components, or generate disconnected parts. More recently, some studies (\cite{zhou2026articraft,lin2026procedura}) have begun to explore agent-based 3D code generation. Although they can produce visually compelling results, they often incur substantial token cost and can't fully capture the underlying geometric constraints and inter-component spatial relationships, resulting in unfaithful reconstructions.

To address this challenge, we propose \textit{\textbf{CADForge}}, an agentic framework for single-view CAD reconstruction generation that converts an input image into a CadQuery program. Rather than generating the entire object monolithically, CADForge decomposes it into individual parts and constructs the model progressively. For each part, CADForge performs explicit geometry reasoning through two stages: \textit{geometry observation} and \textit{geometry programming}. In the first stage, a VLM extracts geometric patterns, proportional cues, and spatial relationships with previously generated parts from the input image. These observations are then translated into mathematical programs that leverage the current CAD state to solve for the key parameters required to construct the part. Based on the obtained specification and parameters, a code generator, assisted by a self-constructed toolkit, generates executable CadQuery code and integrates the new part into the existing model. We also introduce a reviewer agent to assess the result for geometric fidelity and topological validity, diagnose potential errors, and provide targeted feedback for iterative refinement. Throughout this process, CADForge introduces a compact parametric CAD memory to record the current modeling state, preserving cross-component consistency with minimal token overhead. 

We construct an evaluation set covering both part-level and object-level CAD models collected from diverse datasets, and evaluate CADForge against existing methods on it. The results demonstrate that CADForge can more faithfully reconstruct the input image as a parametric CAD model while maintaining competitive token consumption. Our contributions can be summarized as:
\begin{itemize}
    \item We propose CADForge, a multi-stage agentic framework tailored for single-view CAD reconstruction generation that decomposes an object into individual parts and progressively constructs a parametric CAD model through executable CadQuery code.
    \item To accurately capture geometric information from the input image, we introduce an explicit geometry reasoning mechanism that first infers relative geometric relationships among parts and then translates them into key modeling parameters. A compact parametric CAD memory is maintained to ground each reasoning step in the evolving construction state.
    \item We develop a comprehensive benchmark that covers both individual CAD parts and complete multi-part objects from diverse data sources. Experimental results demonstrate that CADForge achieves higher reconstruction fidelity than existing methods.
\end{itemize}

%% file: Sections/2_related_work.tex
\section{Related Work}
\textbf{Parametric CAD Generation.}
Learning-based CAD generation began with efforts to represent construction histories as sequences of sketches and geometric operations (\cite{wu2021deepcad,xu2022skexgen,xu2023hierarchical}). Building on these representations, advances in foundation models and large language models have inspired a growing body of work on conditional CAD generation. These methods translate textual descriptions into editable CAD sequences or executable programs, with some further incorporating explicit geometric controls or intermediate model states to support more complex construction procedures (\cite{khan2024text2cad,li2024cad,li2025cad,zhang2026geocad,guan2026cad,Qi_2026_CVPR}). Alongside this development, another line of research models CAD generation as a reverse-engineering problem, seeking to recover the underlying construction process from geometric observations such as point clouds and multi-view images (\cite{khan2024cad,liu2024point2cad,ma2024draw,rukhovich2025cad,kabisov2026cadreasoner}). More recently, several methods have moved toward unified multimodal frameworks that accommodate heterogeneous inputs, including text, multi-view images, and point clouds, allowing different forms of design intent and geometric evidence to be handled within a shared generation backbone (\cite{xu2024cad,li2025llm4cad,kolodiazhnyi2025cadrille,Pyatov_2026_CVPR}).

Compared with text- or geometry-conditioned settings, single-view CAD reconstruction still faces great challenges. Existing approaches broadly follow two directions: directly predicting CAD command sequences or executable programs (\cite{alam2024gencad,chen2025img2cad,chen2025cadcrafter,doris2026cad}), and recovering structured B-Reps using learned geometric priors and fitting (\cite{li2025caddreamer}). Most of these methods rely on task-specific training with image-CAD datasets that largely consist of simple CAD models. Consequently, their performance often degrades on out-of-distribution images and relatively complicated objects. In contrast, CADForge adopts a training-free agent-based framework, which leverages the visual, geometric, and programming priors of VLMs and generalizes more effectively without specialized fine-tuning.

\textbf{Agentic 3D Modeling.}
Recent advances in large language models and vision-language models have spurred growing interest in agentic 3D modeling, where models plan modeling procedures, interact with 3D software, and refine their outputs through execution and visual feedback. Early systems primarily use LLMs to operate general-purpose 3D software such as Blender, translating natural-language instructions into coordinated modeling actions for object and scene creation (\cite{hu2024scenecraft,sun20253d,zhou2025scenex,lu2026yo}). More work has gone beyond tool control by treating executable programs as structured representations of 3D objects themselves. Compared with direct mesh representations, code makes semantic parts and construction logic explicit, supports editing and reuse, and enables deterministic execution and automated verification (\cite{lu2025ll3m, dai2026meshcoder}). Building on these advantages, some approaches generate compositional assets through specialized programmatic interfaces that explicitly represent parts, spatial relationships, and structural constraints (\cite{barkley2026cadsmith, li2026seek, zhou2026articraft,lin2026procedura}).However, these methods are mostly designed for text-conditioned generation. When applied to image-based reconstruction, they often struggle to recover key geometric patterns and inter-part relationships, and their outputs are not necessarily editable B-Reps. In contrast, CADForge is explicitly designed for single-view CAD reconstruction. It combines a tailored part-decomposition strategy with explicit geometric reasoning to recover geometric structure and spatial relationships from the input image and convert them into an editable parametric CAD model.

%% file: Sections/3_method.tex
\section{Method}
In this section, we present CADForge, which reconstructs a CAD model from an arbitrary image. As illustrated in Fig.~\ref{fig:pipeline}, CADForge first decomposes the complete object into a set of distinct yet interrelated components and then constructs them part by part (\ref{component_decomposing}). For each part, our geometry reasoning module carefully analyzes the input image and formulates mathematical programs to solve for the key geometric parameters (\ref{geometry_reasoning}). Assisted by a self-constructed toolkit, a code generator then synthesizes the corresponding CadQuery code (\ref{code_generation}). To ensure global consistency without introducing excessive context, we maintain a compact parametric CAD memory that records the evolving CAD state and allows agents to retrieve relevant historical parameters when needed (\ref{memory}).
\input{Tex/pipeline}
\subsection{Problem Formulation}

Given a single RGB image $\mathbf{I} \in \mathbb{R}^{H \times W \times 3}$ depicting an object, our goal is to reconstruct the object as a part-aware parametric CAD model $\mathcal{O}=(\mathcal{B},\mathcal{Q})$, where $\mathcal{B}=\{B_1,\ldots,B_N\}$ is a list of B-Rep shapes representing the object’s components, and $\mathcal{Q}=\{q_1,\ldots,q_N\}$ contains the corresponding CadQuery programs. Each program $q_i$ specifies the geometry and spatial placement of the $i$-th component, and is executed by a CAD kernel to produce its shape:
\begin{equation}
B_i = \operatorname{KernelExecute}(q_i).
\end{equation}
Together, the generated components form a coherent reconstruction, while their CadQuery programs preserve parametric editability and construction logic for additional downstream CAD operations.

\subsection{Component Decomposing}
\label{component_decomposing}
To avoid the ambiguity and complexity of synthesizing an entire object in a single step, we first employ a VLM to decompose the object depicted in the input image into an ordered sequence of CAD-meaningful components:
$\mathcal{P}=[(p_1,b_1),\ldots,(p_N,b_N)]$,
where \(p_i\) denotes the label of the \(i\)-th target component (e.g., \texttt{seat\_panel}), and \(b_i\) is a brief description capturing its semantic and structural characteristics. Each \(p_i\) represents either an individual physical part or a coherent subassembly. For occluded or unobserved regions, the VLM combines the available visual evidence with learned object priors to infer plausible missing components and their main forms.

Rather than relying solely on visual appearance, our decomposition reflects practical CAD workflows in both granularity and construction order. Elements with shared geometric structures, such as symmetry or repetition, are treated as a single modeling unit, while the sequence \(\mathcal{P}\) arranges all units according to their dependencies. Specifically, structural parts are constructed earlier to establish an overall geometric scaffold, after which dependent parts are shaped with reference to the existing geometry and placed according to attachment, embedding, or alignment relationships (as visualized in Fig.~\ref{fig:case_decompose}). In this way, CADForge derives a compact, structured blueprint that preserves the object’s underlying construction logic and provides global guidance for progressive CAD reconstruction.

\subsection{Geometry Reasoning}
\label{geometry_reasoning}
A key challenge in single-view CAD reconstruction is recovering geometric structure and spatial relationships from incomplete visual information. Existing methods often fail to infer these cues reliably, leading to geometric inconsistencies in the reconstructed model. To address this, we introduce a two-stage geometry reasoning module before CAD code generation, consisting of geometry observation and geometry programming to extract and formalize CAD-relevant geometric cues.

\subsubsection{Geometry Observation}
For each component \(p_i\), CADForge analyzes the input image \(I\) and selects a compact reference set \(\mathcal{R}_i\) from previously constructed components. The selection is guided primarily by spatial relationships and shape relevance, favoring parts that are spatially related or structurally analogous to $p_i$. Using \(\mathcal{R}_i\) as geometric anchors, the geometry observation module then produces a CAD-oriented specification to guide subsequent shape modeling and component assembly: 
\begin{equation}
g_i = \mathrm{Observe}(I, p_i, b_i, \mathcal{R}_i).
\end{equation}
Concretely, $g_i$ captures both the target component's intrinsic structure and its geometric relationships with $\mathcal{R}_i$. At the intra-component level, \(g_i\) describes its profile, principal orientation, symmetry, and repetition patterns. At the inter-component level, it specifies its pose and scale relative to \(\mathcal{R}_i\), together with the attachment, alignment and support relationships, which may include constraints such as parallelism and perpendicularity, as well as relative offsets between components. Unlike implicit visual cues, this explicit and structured representation situates \(p_i\) relative to the existing geometry and reduces ambiguity, significantly enhancing the consistency of the generated model.

\subsubsection{Geometry Programming}
To bridge the gap between natural-language geometric descriptions and CAD construction, CADForge further introduces Geometry Programming, a module that translates geometric specifications into executable code for computing key modeling parameters. Instead of reasoning about every aspect of a component, this stage focuses on the essential geometric parameters implied by \(g_i\). For each component \(p_i\), CADForge first retrieves the structured parameters \(\Theta_{\mathcal{R}_i}\) of its reference components from the parametric CAD memory (\ref{memory}), and then uses these parameters along with \(g_i\) to synthesize a lightweight Python program \(s_i\), which is executed to obtain the target parameters \(\theta_i\):
\begin{equation}
s_i = \mathrm{ProgramLLM}(g_i, \Theta_{\mathcal{R}_i}), \qquad
\theta_i = \mathrm{Exec}(s_i).    
\end{equation}
The resulting \(\theta_i\) is returned as a structured dictionary that records each parameter’s name, meaning, and numerical value, providing explicit and inspectable geometric guidance for CAD shape modeling. By converting high-level geometric information into precise, construction-ready parameters, our proposed two-stage Geometry Reasoning greatly enhances geometric accuracy and cross-component consistency while reducing the inference burden for downstream code generation. 

\subsection{CadQuery Generation}
\label{code_generation}
\subsubsection{Code Synthesis}
Given the semantic brief \(b_i\), geometric specification \(g_i\), solved parameters \(\theta_i\), and the reference modeling parameters \(\Theta_{\mathcal{R}_i}\), we employ an LLM-based code generator to synthesize an executable CadQuery program for the current component:
\begin{equation}
c_i = \mathrm{CodeLLM}(p_i, b_i, g_i, \theta_i, \Theta_{\mathcal{R}_i}).
\end{equation}
Here, \(c_i\) denotes the CadQuery program that constructs \(p_i\), which is compiled together with previously generated programs to produce the updated CAD model. Since Geometry Reasoning has already parameterized and resolved the key geometric relationships, the code generator can devote more modeling capacity to generating appropriate CadQuery operations rather than determining spatial configuration, which leads to more stable code synthesis and more faithful CAD geometry. If the generated program fails to execute or produces CAD output that does not pass the validity checks, a code repairer revises the CadQuery code based on the associated error message.

After successful execution, CADForge renders the updated model from two complementary views, with new components highlighted in different colors, and concatenates them with the input image for review. The Review Agent then checks whether the generated component matches the intended shape, scale, and position. If a failure is detected, it attributes the error to either geometry reasoning or code synthesis and routes the revision back to the corresponding stage with targeted feedback.

\subsubsection{Failure-Guided Toolkit Self-Construction}
Although CadQuery provides a powerful procedural framework for CAD modeling, synthesizing reliable CadQuery code remains challenging due to its stateful, selector-driven modeling paradigm and strict geometric validity requirements. Even syntactically correct programs may still fail or produce invalid geometry. To further improve robustness and prevent recurring implementation errors, we propose a failure-guided mechanism for automatically constructing a reusable toolkit.

Specifically, we construct a separate set from the Fusion 360 dataset (\cite{willis2021fusion}) and run our pipeline on the corresponding rendered images, recording complete generation traces that include intermediate observations, geometry programs, synthesized code, execution results, and review feedback. From these traces, we mine two complementary types of failure cases. The first consists of runs with repeated code execution failures, which reveal CadQuery operations that LLMs struggle to implement reliably. The second includes geometrically inaccurate generations, identified by ranking all cases based on the Chamfer Distance and selecting the worst 30\% as low-quality cases.
An LLM then analyzes these cases together with their generation traces to summarize recurring failure patterns and identify the underlying CAD operations responsible for them. Based on these summaries, we prompt the LLM to encapsulate frequently used yet error-prone modeling operations into robust, reusable helper functions, such as those for solid composition and hole construction. Only helpers that have been verified as effective are incorporated into the toolkit. As new failure traces are collected, the same process can be repeated to expand the toolkit incrementally.
During code synthesis, the code generator can directly invoke relevant helpers in the toolkit for fragile operations instead of reimplementing them from scratch, significantly improving reliability and efficiency. 

\subsection{Parametric CAD Memory}
\label{memory}
Since CADForge constructs an object component by component, each generation step may depend on the dimensions, spatial placement, and attachment interfaces of previously generated components. Passing the full construction history to the LLM would introduce substantial redundancy and token overhead. To avoid this, CADForge maintains a lightweight parametric CAD memory, implemented as a nested dictionary in which each accepted component is associated with its important modeling parameters. Specifically, when generating the CadQuery program for \(p_i\), CADForge further abstracts the underlying modeling operations into a structured parameter dictionary \(\Theta_i=\{(k_j,v_j)\}_{j=1}^{m_i}\), where each key \(k_j\) is a concise parameter name (e.g., \texttt{full\_width}) and \(v_j\) stores its corresponding numerical value. Once \(p_i\) is accepted, \(\Theta_i\) is added to the memory as follows:
\begin{equation}
\mathcal{M}_i = \mathcal{M}_{i-1} \cup ({p_i \mapsto \Theta_i}).
\end{equation}
For each component, CADForge queries $\mathcal{M}_i$ with the labels of its reference components and retrieves the corresponding parameter entries to guide Geometry Reasoning and Code Synthesis, thereby grounding the generation in the existing geometry while avoiding redundant information.

%% file: Tex/pipeline.tex
\begin{figure}[t]
    \centering
    \includegraphics[width=\linewidth]{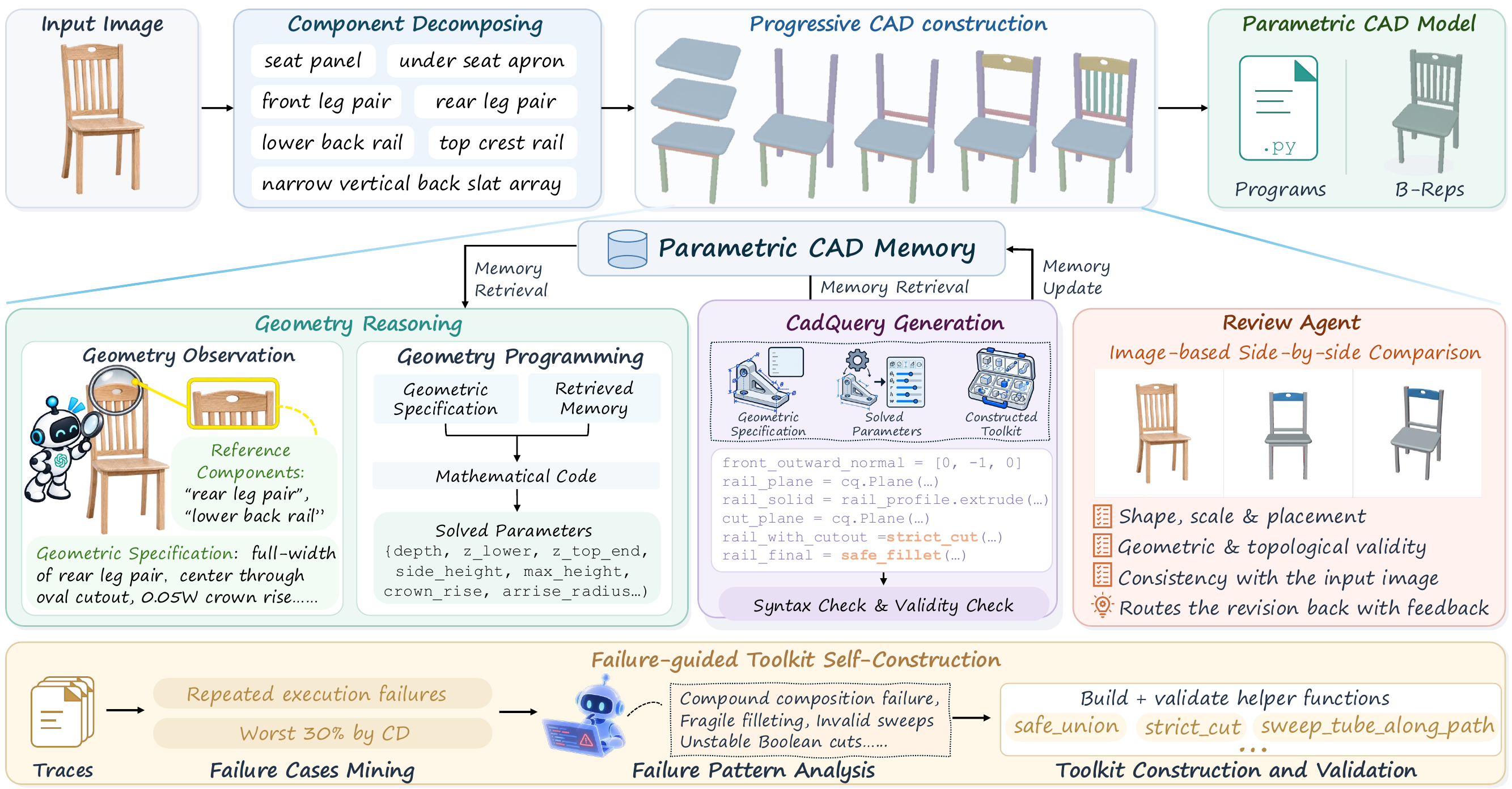}
    \caption{
        \textbf{Overview of CADForge.}
        Given a single image, CADForge decomposes the object into several components and constructs them progressively. For each component, geometry observation identifies reference components and CAD-relevant constraints, while geometry programming translates them into key modeling parameters. A code generator then synthesizes executable CadQuery code, and a review agent evaluates the result and provides targeted feedback. Throughout the process, a compact parametric CAD memory maintains essential CAD state for on-demand retrieval, while a failure-guided toolkit offers reusable helpers for robust code generation.
    }
    \label{fig:pipeline}
\vspace{-9pt}
\end{figure}

%% file: Sections/4_experiments.tex
\section{Experiments}
\subsection{Settings}
\textbf{Evaluation Set}. 
To comprehensively assess our method, we construct an evaluation set of 200 samples. Among them, 170 samples are collected from Objaverse (\cite{deitke2023objaverse}), Fusion 360 (\cite{willis2021fusion}), and 3D-Future (\cite{fu20213d}), covering single-part objects, multi-part objects, everyday items, and furniture. These samples are used to evaluate geometric metrics. Unless otherwise specified, all subsequent geometry-related evaluations and ablation studies are conducted on this  subset. To further demonstrate the out-of-distribution (OOD) generalization capability of our method, we additionally select 30 web images for perceptual evaluation. The entire evaluation set is strictly disjoint from the data used for toolkit construction. 

\textbf{Baselines}. Focusing on the single-view CAD reconstruction task, we primarily evaluate mainstream vision-language models (VLMs) and current open-sourced agentic approaches. The evaluated VLMs include MiniMax-M3 (\cite{lai2026minimax}), Kimi-K3 (\cite{team2026kimi}), GPT-5.5, and GPT-5.6-Sol, while the agentic baselines consist of CAD-Coder (\cite{doris2026cad}), Articraft (\cite{zhou2026articraft}), and the concurrent work Procedura (\cite{lin2026procedura}). 
\input{Tables/main_table}
\input{Tables/ablation_component}

\textbf{Metrics}.
To evaluate the geometric reconstruction quality of different methods, we adopt Chamfer Distance (CD-L1) (\cite{mescheder2019occupancy}), Hausdorff Distance (HD) (\cite{huttenlocher1993comparing}), F-Score (\cite{tatarchenko2019single}), Silhouette Intersection over Union (SIOU) (\cite{wiles2019learning}), and Normal Consistency (NC) (\cite{mescheder2019occupancy}) as our primary metrics. For agentic methods, we additionally report the average total number of tokens consumed to synthesize a CAD model (AvgT), including both input and output tokens. To assess the ability to generate valid CAD models, we further introduce the CAD Program Validity Rate (CPVR), defined as the proportion of all construction attempts that successfully produce a valid CAD model. Besides these objective metrics, we further conduct VLM-based and human preference evaluations, quantified by the preference rates at the top three ranks. More details are provided in ~\ref{appendix:metrics}.

\textbf{Implementation Details}. 
All experiments were conducted on the same server and reproduced according to the official documentation or official open-source repositories. For VLMs such as GPT-5.5 and MiniMax-M3, 
we repeatedly query the model until it generates executable CadQuery code.
All agentic methods use GPT-5.5 as the backbone LLM/VLM. For CADForge, each generation step allows at most five code-repair attempts and three failed review attempts. For models generated by different methods, we focus on their geometric structures. Please refer to ~\ref{sup:imple} for more specifics.

\subsection{Quantitative Comparison}
\textbf{Geometric Evaluation}. As shown in Tab.~\ref{tab:comparison}, CADForge achieves the best performance across all geometric metrics while maintaining a strong CPVR. It reduces CD-L1 to 0.01580, improving over Procedura by 8.35\%, and achieves 0.52104 on the more stringent F@0.01 metric, demonstrating better recovery of fine-grained structures. Improvements in HD, SIOU, and NC further confirm its overall geometric fidelity, while its competitive CPVR indicates reliable generation of valid B-reps. Moreover, CADForge consumes only 81.6k tokens on average, 75.4\% fewer than Procedura, achieving a favorable balance among geometric quality, CAD validity, and token efficiency.

\input{Tables/ablation_toolkit}

\input{Tex/preference}
\textbf{Preference Evaluation}.
To complement geometric metrics with perceptual assessment, we further conduct a VLM-based preference evaluation and report the resulting Rank-\(k\) (\(k=1,2,3\)) preference rates in Fig.~\ref{fig:preference-results}. Specifically, for each input, we render the outputs of the five compared methods from a common viewpoint and arrange them alongside the input image in a single composite image. Gemini-3.1-Pro-Preview then serves as the VLM judge and selects the top three outputs based on an overall assessment of shape similarity, proportional consistency, and geometric fidelity. CADForge achieves the highest Rank-1 preference rate, indicating its strong performance in structural plausibility, geometric accuracy, and perceptual quality. A human user study is provided in \ref{user_study}.

\subsection{Qualitative Comparison}
\input{Tex/case}
Fig. ~\ref{fig:case} presents qualitative comparisons across different methods. CADForge consistently achieves higher reconstruction fidelity and better geometric consistency, particularly for objects with repetitive structures and fine-grained geometric details. For example, on the L-shaped bracket in the first row, CADForge accurately recovers the number and arrangement of holes, whereas competing methods often miss, distort, or misplace them. Similar structural errors can be observed in the frame and table examples, where other methods introduce distorted joints, incorrect supports, or inconsistent component connections. The differences become more pronounced for complex storage furniture: competing methods frequently produce misaligned partitions, merged components, or incorrect shelf layouts, while CADForge preserves the intended structure. Overall, CADForge produces cleaner, more coherent, and structurally faithful CAD reconstructions across a diverse range of objects.

\subsection{Ablation Studies}
\textbf{Ablation on Framework Components}. We conduct ablation studies on the three key modules of CADForge: component decomposing, geometry reasoning, and the review agent, with quantitative results reported in Tab.~\ref{tab:ablation_component} and qualitative comparisons presented in Fig.~\ref{fig:ablation_case}. Experimental results show that removing any module degrades performance, validating the coherence of our unified design and the effectiveness of each stage. As shown in Fig.~\ref{fig:ablation_case}, when component decomposition is replaced with step-wise component selection based on the current partial result, the reconstruction is prone to missing components due to the lack of global planning. Similarly, removing geometry reasoning noticeably compromises geometric accuracy, causing distortions such as incorrect aspect ratios.  Without the review agent, the model is less capable of identifying and correcting existing mistakes, often leading to error accumulation over successive generation steps. We further provide an ablation study on geometry programming in the supplementary material.

\textbf{Ablation on Toolkit}. To examine the impact of the failure-guided self-constructed toolkit, we remove it from CADForge for ablation in Tab.~\ref{tab:toolkit_ablation}. Without it, both the CPVR and Average Repair Count deteriorate, suggesting that the code generator more frequently produces non-executable code and requires additional repair iterations, consequently increasing the Average Token consumption. Moreover, the absence of modeling helpers can limit the code generator’s ability to construct some shapes or structures, degrading metrics such as HD. As evidenced in the bottom-right example of Fig.~\ref{fig:ablation_case}, despite multiple rounds of code repair and review, the code generator without toolkit still fails to accurately model the cylindrical component, resulting in a clear mismatch with the input image.

\textbf{Ablation on Parametric CAD Memory}.
We further conduct an ablation study of the parametric CAD memory on the 3D-FUTURE subset (Tab.~\ref{tab:memory_ablation}). Since CADForge generates models component by component, each new component has to depend on information from previous ones. Thus, completely removing memory without any historical context is not a meaningful ablation. For the\input{Tables/ablation_memory}variant without our proposed memory, we follow prior methods and provide the full generation histories of the reference components. Besides the geometric metrics, we additionally report the number of tokens consumed per component (Tok./Comp.). While both variants achieve comparable generation performance, our memory substantially reduces token consumption, demonstrating that it effectively compresses long generation histories while preserving the needed geometric information.

\textbf{Ablation on VLM Backbones}. To assess the generalizability of our framework, we replace the backbone of CADForge with GPT-5.6-Sol and Gemini-3.1-Pro-Preview and benchmark it under different model configurations. The results in Tab.~\ref{tab:backbone} indicate that CADForge maintains strong results across different backbone models, demonstrating that its effectiveness is not tied to a specific foundation model. Importantly, CADForge with GPT-5.6-Sol delivers the best overall performance, implying that advances in foundation model capability can be effectively translated into further gains within our agentic framework. We additionally provide visual comparisons in Fig.~\ref{fig:case_backbones}.
\input{Tables/ablation_backbone}
\input{Tex/ablation_case}

%% file: Tables/main_table.tex
\begin{table}[t]
\caption{\textbf{Quantitative comparison with baselines}.
We report CAD Program Validity Rate (CPVR), CD-L1, HD, F-scores,
Silhouette Intersection over Union (SIOU), Normal Consistency (NC). For agentic methods, we report Average Token Consumption (AvgT). The best results are in \textbf{bold}.}
\label{tab:comparison}
\begin{center}
\resizebox{\linewidth}{!}{%
\begin{tabular}{lcccccccc}
\toprule
Method
& AvgT$\downarrow$
& CD-L1$\downarrow$
& HD$\downarrow$
& F@0.05$\uparrow$
& F@0.01$\uparrow$
& SIOU$\uparrow$
& NC$\uparrow$
& CPVR$\uparrow$ \\
\midrule
MiniMax-M3
& -- & 0.02465 & 0.13204 & 0.86503 & 0.42338 & 0.66092 & 0.67174 & 80.40\% \\
Kimi-K3
& -- & 0.01962 & 0.09968 & 0.92051 & 0.44060 & 0.71130 & 0.69883 & 78.90\% \\
GPT-5.5
& -- & 0.01867 & 0.10329 & 0.92096 & 0.47000 & 0.71382 & 0.68976 & 83.49\% \\
GPT-5.6-Sol
& -- & 0.01735 & 0.09427 & 0.93256 & 0.49162 & 0.73341 & 0.71575 & 85.97\% \\
\midrule
CAD-Coder
& -- & 0.04500 & 0.20964 & 0.66808 & 0.26644 & 0.55691 & 0.57553 & \textbf{94.59\%} \\
Articraft
& 236548 & 0.01759 & 0.10436 & 0.93115 & 0.49598 & 0.73303 & 0.68912 & 62.15\% \\
Procedura
& 331804 & 0.01724 & 0.10032 & 0.93997 & 0.46771 & 0.73180 & 0.69902 & 86.84\% \\
\textbf{CADForge (Ours)}
& \textbf{81613}
& \textbf{0.01580}
& \textbf{0.08749}
& \textbf{0.94616}
& \textbf{0.52104}
& \textbf{0.75300}
& \textbf{0.72332}
& 93.81\% \\
\bottomrule
\end{tabular}%
}
\end{center}
\vspace{-10pt}
\end{table}

%% file: Tables/ablation_component.tex
\begin{table}[t]
\centering
\caption{\textbf{Ablation study of three key CADForge components}.}
\vspace{-6pt}
\begin{center}
\label{tab:ablation_component}
\resizebox{\linewidth}{!}{%
\begin{tabular}{lcccccc}
\toprule
Method & CD-L1$\downarrow$ & HD$\downarrow$ & F@0.05$\uparrow$ & F@0.01$\uparrow$ & SIOU$\uparrow$ & NC$\uparrow$ \\
\midrule
w/o \ component decomposing & 0.01769 & 0.10338 & 0.92886 & 0.48224 & 0.73184 & 0.70218 \\
w/o \ geometry reasoning    & 0.01834 & 0.10554 & 0.92253 & 0.47624 & 0.72935 & 0.70094 \\
w/o \ review agent               & 0.01941 & 0.10971 & 0.91455 & 0.46345 & 0.71512 & 0.69510 \\
\midrule
\textbf{CADForge} & \textbf{0.01580} & \textbf{0.08749} & \textbf{0.94616} & \textbf{0.52104} & \textbf{0.75300} & \textbf{0.72332} \\
\bottomrule
\end{tabular}%
}
\end{center}
\vspace{-16pt}
\end{table}

%% file: Tables/ablation_toolkit.tex
\begin{table}[t]
\centering
\caption{\textbf{Ablation study of the CADForge toolkit}. Average Repair Count (Avg.Repair Count) represents the average number of repair iterations required to fix syntax errors in the generated code.}
\vspace{6pt}
\label{tab:toolkit_ablation}
\resizebox{0.95\linewidth}{!}{%
\begin{tabular}{lccccccc}
\toprule
Method
& Avg. Tokens$\downarrow$
& CPVR$\uparrow$ 
& \shortstack{Avg. Repair Count}$\downarrow$
& CD-L1$\downarrow$
& HD$\downarrow$
& F@0.05$\uparrow$
& \shortstack{NC}$\uparrow$ \\
\midrule
w/o toolkit
& 85943
& 91.08\%
& 1.68750
& 0.01729
& 0.09646
& 0.93345
& 0.70758 \\
\midrule
\textbf{CADForge}
& \textbf{81613}
& \textbf{93.81\%}
& \textbf{1.29167}
& \textbf{0.01580}
& \textbf{0.08749}
& \textbf{0.94616}
& \textbf{0.72332} \\
\bottomrule
\end{tabular}%
}
\vspace{-12.6pt}
\end{table}

%% file: Tex/preference.tex
\begin{wrapfigure}{r}{0.42\textwidth}
  \vspace{-0.8\baselineskip}
  \centering
  \includegraphics[width=\linewidth]{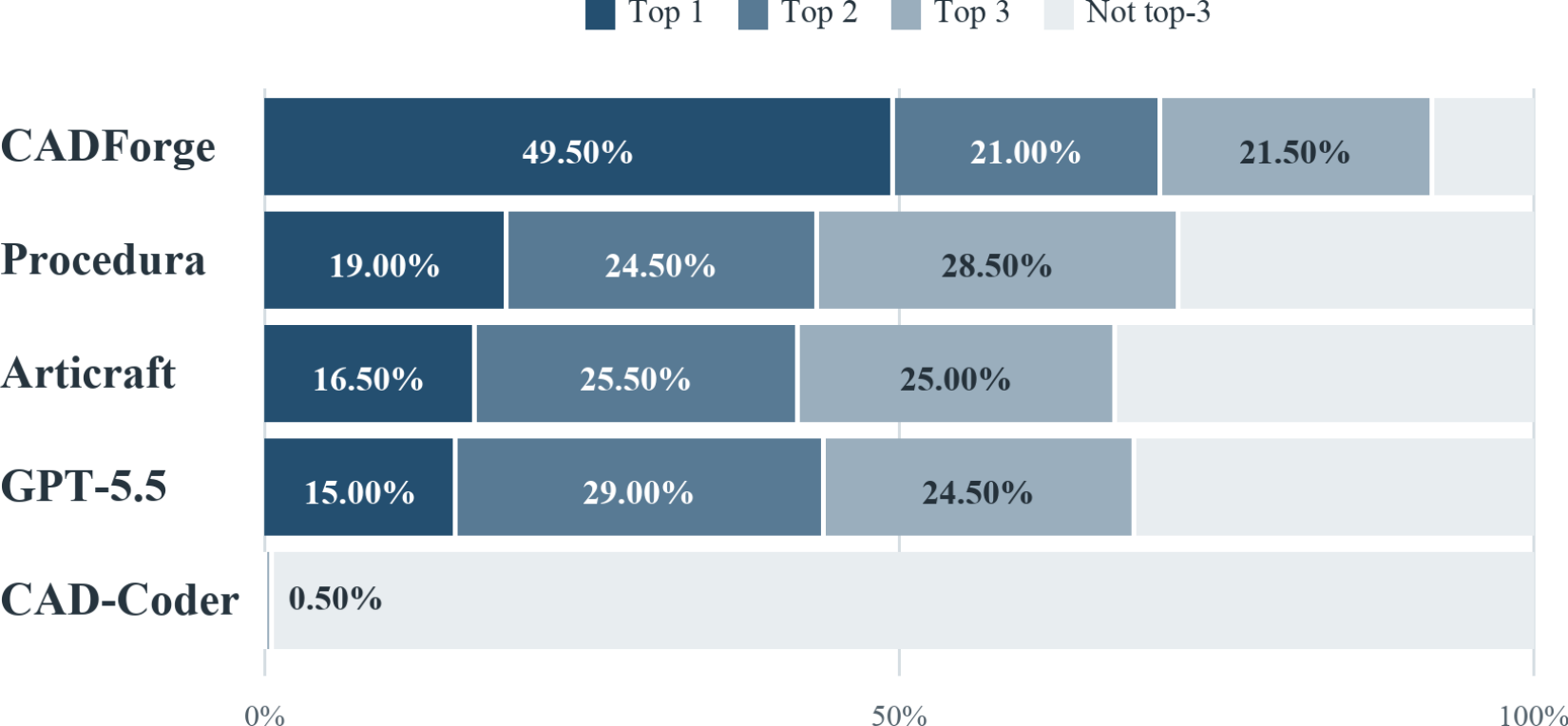}
  \caption{VLM-based preference evaluation across different methods.}
  \label{fig:preference-results}
  \vspace{-0.5\baselineskip}
\end{wrapfigure}

%% file: Tex/case.tex
\begin{figure}[t]
    \centering
    \includegraphics[width=\linewidth]{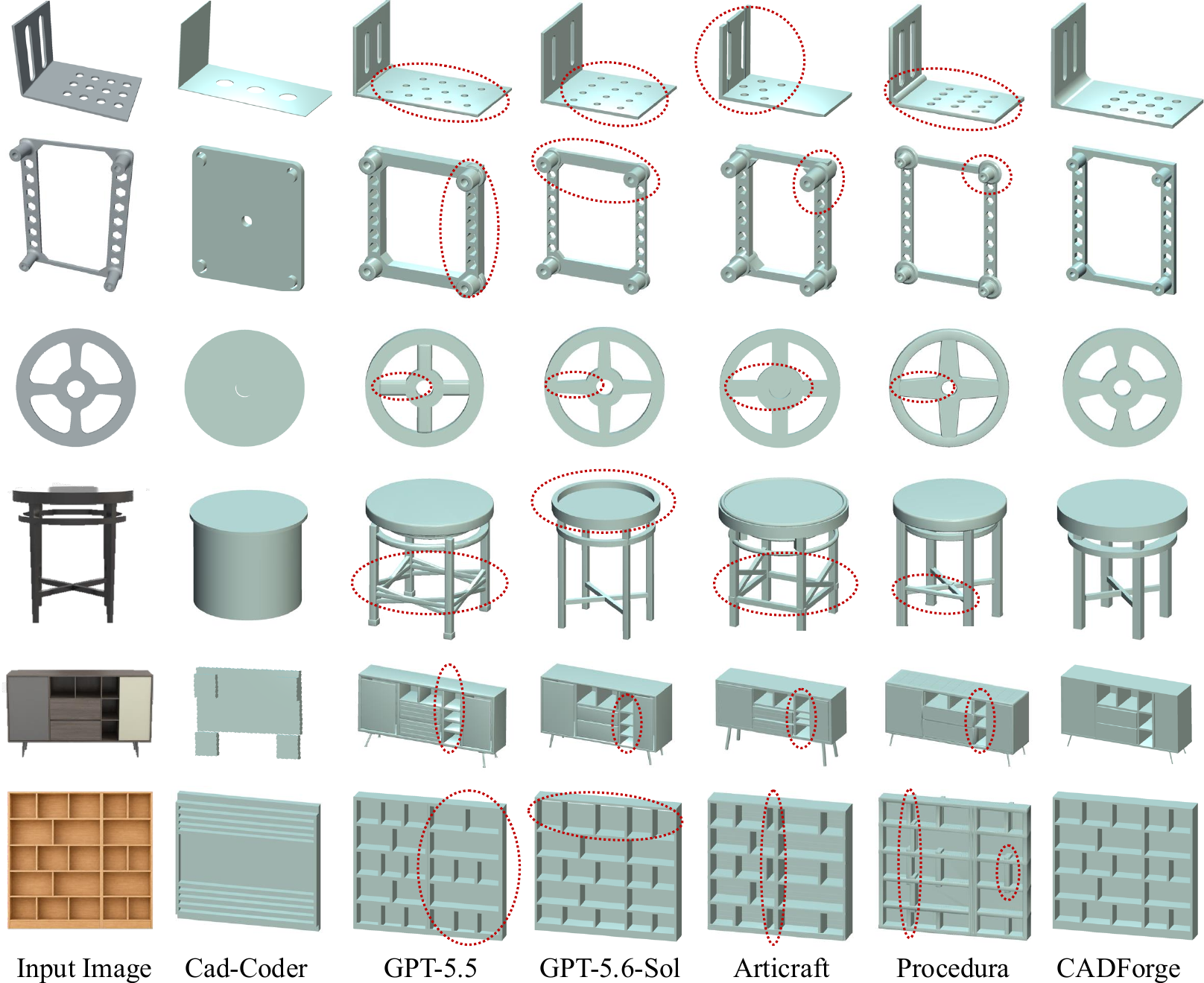}
    \caption{\textbf{Qualitative comparison of results generated by different methods}. Red circles highlight regions with geometric inconsistencies relative to the input images.
    }
    \label{fig:case}
\vspace{-12pt}
\end{figure}

%% file: Tables/ablation_memory.tex
\begin{wraptable}{r}{0.43\linewidth}
\vspace{-16pt}
\caption{\textbf{Ablation on the CAD memory}.}
\vspace{3pt}
\label{tab:memory_ablation}
\resizebox{\linewidth}{!}{%
\begin{tabular}{lccc}
\toprule
 & Tok./Comp.$\downarrow$ & CD-L1$\downarrow$ & HD$\downarrow$ \\
\midrule 
w/o Memory & 45696 & 0.01595 & 0.10137 \\
w/ Memory   & 27225 & 0.01601 & 0.09332 \\
\bottomrule
\end{tabular}%
}
\vspace{-6pt}
\end{wraptable}

%% file: Tables/ablation_backbone.tex
\begin{table}[t]
\caption{\textbf{Performance of CADForge with different backbone models}.}
\label{tab:backbone}
\begin{center}
\resizebox{0.9\linewidth}{!}{%
\begin{tabular}{lcccccc}
\toprule
Backbone &
CD-L1$\downarrow$
& HD$\downarrow$
& F@0.05$\uparrow$
& F@0.01$\uparrow$
& SIOU $\uparrow$
& NC $\uparrow$ \\

\midrule

GPT-5.6-Sol
& 0.01539
& 0.08740
& 0.94076
& 0.56187
& 0.76068
& 0.73160 \\

Gemini-3.1-Pro-Preview
& 0.01620
& 0.09299
& 0.93768
& 0.52581
& 0.74610
& 0.71549 \\

\textbf{GPT-5.5~(Ours)}
& 0.01580
& 0.08749
& 0.94616
& 0.52104
& 0.75300
& 0.72332 \\

\bottomrule
\end{tabular}%
}
\end{center}
\vspace{-12pt}
\end{table}

%% file: Tex/ablation_case.tex
\begin{figure}[t]
    \centering
    \includegraphics[width=0.96\linewidth]{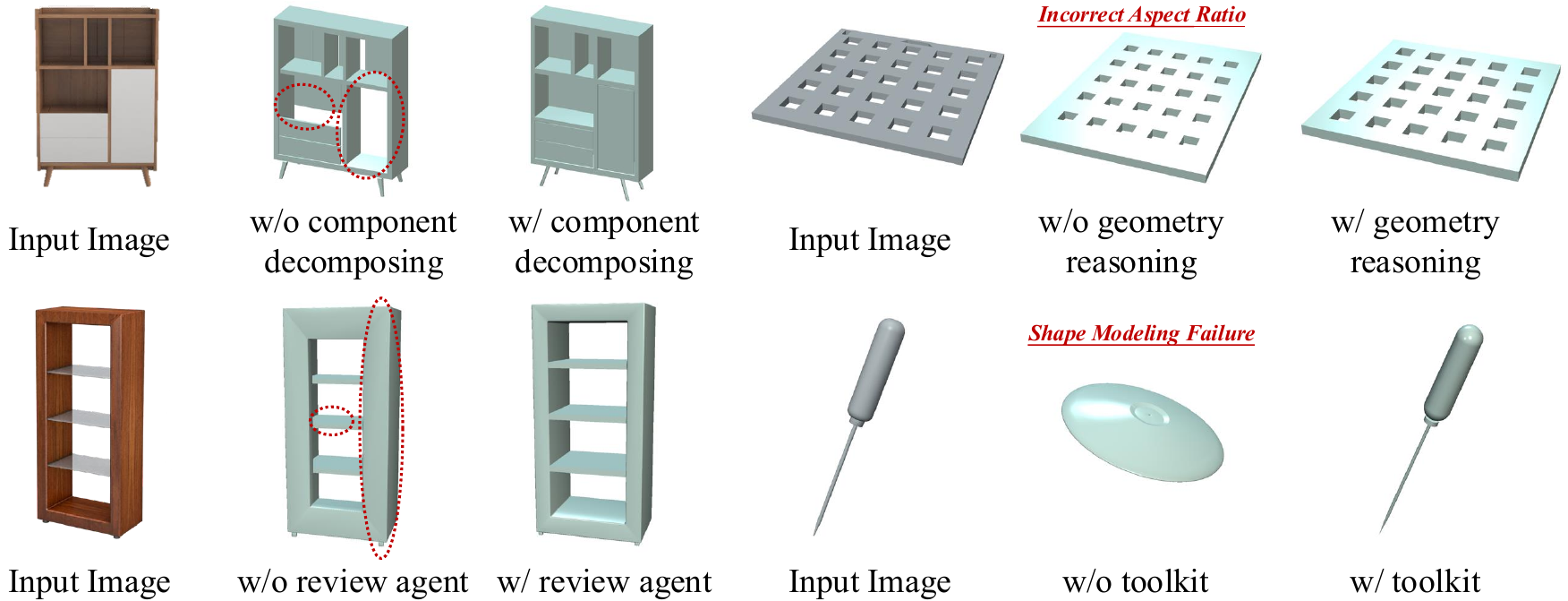}
    \caption{\textbf{Qualitative comparison of the ablation studies}, illustrating the effectiveness of component decomposition, geometry reasoning, the review agent, and the self-constructed toolkit.}
    \label{fig:ablation_case}
\vspace{-16pt}
\end{figure}

%% file: Sections/5_conclusion.tex
\section{Conclusion}
We present CADForge, a training-free agentic framework for single-view CAD reconstruction. CADForge progressively reconstructs CAD models by decomposing objects into components, explicitly reasoning about their geometry, and iteratively refining the generated programs, while improving robustness and efficiency through a failure-guided toolkit and compact parametric memory.  Experiments show that CADForge achieves higher reconstruction fidelity with lower token consumption than existing baselines. Overall, our results demonstrate the effectiveness of CADForge for accurate and efficient parametric CAD reconstruction from a single-view image.

%% file: Sections/X_supplementary.tex
\clearpage
\section{AI Use Statement}
Large language models and AI tools were used for two purposes. First, they assisted with language polishing, including wording, grammar, and sentence structure. All AI-assisted revisions were carefully reviewed and proofread by the authors to ensure factual correctness, preserve the intended meaning, and avoid hallucinated or unsupported statements. Second, as CADForge is an agentic framework, VLMs were used as components of the proposed pipeline and for perceptual assessment, as described in the paper. All such uses followed human-defined procedures and were subject to human supervision and verification. The research design, experimental protocol, result interpretation, and scientific conclusions were determined by the authors. The authors take full responsibility for the final content of this work, including text, claims and conclusions.

\section{Experimental Details}
\subsection{Evaluation Set}
To assess single-view CAD reconstruction across diverse object categories, we curate an evaluation set of 200 samples, including 170 objects from Objaverse (\cite{deitke2023objaverse}), Fusion 360 (\cite{willis2021fusion}), and 3D-Future (\cite{fu20213d}), as well as 30 web images. The 170 samples span simple and complex CAD parts, everyday objects, and furniture, and are used for geometric evaluation. For Objaverse and Fusion 360, we follow the data-processing protocol of CADBench (\cite{doris2026cadbench}). Each object is rendered from one of several representative viewpoints rather than a fixed canonical view. The 30 web images are collected to assess out-of-distribution generalization through qualitative comparison and perceptual assessment. Examples are shown in Fig.~\ref{fig:dataset_presentation}.

\section{Baselines}
We primarily compare CADForge with two categories of methods that are directly executable under our evaluation setting: (1) general-purpose VLMs capable of generating CadQuery programs from images, including MiniMax-M3, Kimi-K3, GPT-5.5, and GPT-5.6-Sol; and (2) recent CAD-related agentic or programmatic 3D modeling methods, including CAD-Coder, Articraft, and Procedura. We reproduce all open-sourced methods following their official implementations and use the same backbone model whenever applicable.

Several recent single-view CAD reconstruction methods are closely related to our task. However, some of them such as Img2CADSeq (\cite{tan2026img2cadseq}), CADCrafter (\cite{chen2025cadcrafter}), and Img2CAD (\cite{chen2025img2cad}) do not provide publicly available code or inference checkpoints, while other works, such as CADDreamer (\cite{li2025caddreamer}), adopt a different setting from ours. Therefore, we do not include them in the main quantitative comparison. 

\subsection{Metrics}
\label{appendix:metrics}

\subsubsection{Metric Computation}
We first put each generated output and its corresponding ground truth in the same canonical object-centric coordinate frame, where the \(x\)-axis points to the object’s right, the \(y\)-axis points from the object’s canonical front toward its back, and the \(z\)-axis points upward. We then translate both to the origin based on their geometric centers and apply scale normalization. Then we uniformly sample \(N_s=50{,}000\) points from the surfaces of both the predicted model and the ground-truth model. Let $
P=\{(\mathbf{p}_i,\mathbf{n}^{P}_i)\}_{i=1}^{N_s},
G=\{(\mathbf{g}_j,\mathbf{n}^{G}_j)\}_{j=1}^{N_s}$
denote the sampled surface points and their corresponding face normals. For a point \(\mathbf{x}\) and a point set \(Y\), we define its nearest-neighbor distance as
\begin{equation}
d(\mathbf{x},Y)=\min_{\mathbf{y}\in Y}\|\mathbf{x}-\mathbf{y}\|_2,
\end{equation}
and its nearest-neighbor index as
\begin{equation}
\pi_Y(\mathbf{x})
=
\arg\min_{\mathbf{y}\in Y}\|\mathbf{x}-\mathbf{y}\|_2.    
\end{equation}

\input{Tex/dataset_presentation}

\textbf{Chamfer Distance (CD-L1).} 
We adopt the L1 variant of Chamfer Distance and measure the average bidirectional discrepancy between the predicted and ground-truth surfaces:
\[
\operatorname{CD}(P,G)
=
\frac{1}{2}
\left[
\frac{1}{|P|}
\sum_{\mathbf{p}\in P}d(\mathbf{p},G)
+
\frac{1}{|G|}
\sum_{\mathbf{g}\in G}d(\mathbf{g},P)
\right],
\]
where \(d(\mathbf{x},Y)=\min_{\mathbf{y}\in Y}\|\mathbf{x}-\mathbf{y}\|_2\).
The first term measures the distance from the generated surface to the ground truth, while the second measures how well the generated model covers the ground-truth surface. Lower values indicate greater geometric similarity.

\textbf{Hausdorff Distance (HD).} While CD measures the average surface discrepancy, Hausdorff Distance captures the largest bidirectional deviation:
\begin{equation}
\operatorname{HD}(P,G)
=
\max
\left\{
\max_{\mathbf{p}\in P}d(\mathbf{p},G),
\;
\max_{\mathbf{g}\in G}d(\mathbf{g},P)
\right\}.    
\end{equation}
HD is particularly sensitive to severe local errors, such as an excessively long component, a detached part, or a missing region. Since it is computed from finite surface samples, the reported value is a sampled approximation of the exact mesh-to-mesh Hausdorff Distance. Lower values are better.

\textbf{F-score.} Given a normalized distance threshold \(\tau\), we compute geometric precision and recall as
\begin{equation}
\operatorname{Prec}_{\tau}(P,G)
=
\frac{1}{|P|}
\sum_{\mathbf{p}\in P}
\mathbb{I}\!\left[d(\mathbf{p},G)<\tau\right],
\quad
\operatorname{Rec}_{\tau}(P,G)
=
\frac{1}{|G|}
\sum_{\mathbf{g}\in G}
\mathbb{I}\!\left[d(\mathbf{g},P)<\tau\right],
\end{equation}
where \(\mathbb{I}[\cdot]\) is the indicator function. Their harmonic mean gives
\begin{equation}
\operatorname{F}_{\tau}
=
\frac{
2\operatorname{Prec}_{\tau}\operatorname{Rec}_{\tau}
}{
\operatorname{Prec}_{\tau}+\operatorname{Rec}_{\tau}
}.
\end{equation}

When both precision and recall are zero, the F-score is set to zero. We report \(F@0.01\) and \(F@0.05\), with thresholds set to \(1\%\) and \(5\%\) of the normalized bounding-box diagonal, respectively. \(F@0.01\) emphasizes fine-scale geometric accuracy, whereas \(F@0.05\) reflects structural agreement under a more tolerant threshold.

\textbf{Silhouette Intersection over Union (S-IoU).}
To evaluate projected shape consistency, we sample \(N_v=100{,}000\) surface points from each aligned mesh and project them onto a shared image plane from four predefined orthographic views:
\begin{equation}
\mathcal{V}
=
\{ \text{front}, \text{side}, \text{top}, \text{oblique} \}.
\end{equation}
The front, side, and top views correspond to projections onto the \(xz\), \(yz\), and \(xy\) planes, respectively. The oblique view is obtained by applying a \(45^\circ\) yaw rotation before projection. The projected points are rasterized into binary occupancy masks at a resolution of \(384\times384\). We apply the same rasterization and mask-processing procedure to both the prediction and the ground truth.

Let \(S^{P}_{v}\) and \(S^{G}_{v}\) denote the predicted and ground-truth silhouette masks under view \(v\). The silhouette IoU for that view is
\begin{equation}
\operatorname{IoU}_{v}
=
\frac{
|S^{P}_{v}\cap S^{G}_{v}|
}{
|S^{P}_{v}\cup S^{G}_{v}|
}.
\end{equation}
The final S-IoU is the unweighted mean over all evaluation views:
\begin{equation}
\operatorname{S\text{-}IoU}(P,G)
=
\frac{1}{|\mathcal{V}|}
\sum_{v\in\mathcal{V}}
\operatorname{IoU}_{v}.
\end{equation}

Unlike point-wise surface distances, S-IoU directly evaluates consistency in the projected outer shape and visible spatial occupancy. It is therefore sensitive to differences in global proportions, component coverage, and projected layout. Higher values indicate better silhouette fidelity.

\textbf{Normal Consistency}.
To measure local surface orientation, we compare the normal at each sampled point with the normal of its nearest neighbor on the other surface. Bidirectional normal consistency is computed as
\begin{equation}
\operatorname{NC}(P,G)
=
\frac{1}{2}
\left[
\frac{1}{|P|}
\sum_{\mathbf{p}_i\in P}
\left|
\mathbf{n}^{P}_i
\cdot
\mathbf{n}^{G}_{\pi_G(\mathbf{p}_i)}
\right|
+
\frac{1}{|G|}
\sum_{\mathbf{g}_j\in G}
\left|
\mathbf{n}^{G}_j
\cdot
\mathbf{n}^{P}_{\pi_P(\mathbf{g}_j)}
\right|
\right].
\end{equation}
The absolute dot product makes the metric robust to inconsistent face winding introduced during CAD-to-mesh conversion, treating opposite normal signs as equivalent. Normal Consistency ranges from \(0\) to \(1\), with higher values indicating more consistent local surface orientation.

\textbf{CAD Program Validity Rate}. For CAD generation, reliably producing valid B-reps is particularly important. Final generation success alone, however, does not fully characterize execution stability for agentic methods, since failed programs can often be iteratively repaired or regenerated until a valid result is eventually obtained. Two methods may therefore achieve the same final success rate while requiring substantially different numbers of failed executions and repair attempts. To capture this behavior, we additionally record all code execution attempts during inference and report the CAD Program Validity Rate (CPVR) as a complementary measure of execution stability.
Because different methods adopt inherently different generation and repair workflows, defining a strictly workflow-independent notion of an “attempt” is difficult. We therefore follow each method’s native execution procedure rather than artificially imposing a unified repair protocol. Specifically, for VLMs and CAD-Coder, each generated program is counted as one execution attempt; for CADForge and Procedura, each component-level construction attempt, including the initial generation and subsequent repair attempts, is counted separately; and for Articraft, each compilation attempt along its generation trajectory is counted as one attempt. We define
\begin{equation}
    \mathrm{CPVR}
    =
    \frac{N_{\mathrm{success}}}{N_{\mathrm{exec}}}
    =
    1-\frac{N_{\mathrm{fail}}}{N_{\mathrm{exec}}},
\end{equation}
where an attempt is considered successful if it executes correctly and produces a valid geometry. If a model can consistently produce valid CAD programs at each construction step, this suggests stronger intrinsic CAD modeling capability. Thus, a higher CPVR indicates more stable code generation and less dependence on iterative code repair.

\subsubsection{Preference Evaluation Criteria}
\label{criteria}
The aforementioned metrics, such as CD-L1, HD, and SIOU, primarily evaluate generated CAD models from a geometric perspective by measuring geometric distances, shape discrepancies, and spatial overlap. However, they do not fully capture perceptual quality or human preferences regarding the overall appearance and structural plausibility of the generated models. To complement these geometric metrics, we conduct both VLM-based and human preference evaluations. For these evaluations, we compare CADForge with CAD-Coder, two other agentic methods, Procedura and Articraft, and GPT-5.5, which serves as the backbone model of our framework. Both evaluations follow a shared set of criteria covering overall fidelity, global geometry, component-level structure, local CAD features, and potential structural defects, as detailed below.

\begin{tcolorbox}[
    title={Prompt for Preference Evaluation},
    colback=gray!5,
    colframe=gray!60,
    boxrule=0.6pt,
    arc=2pt,
    left=6pt,
    right=6pt,
    top=6pt,
    bottom=6pt,
    fonttitle=\bfseries,
    breakable
]
\label{criteria}
\textbf{System Prompt.}
You are a rigorous and impartial evaluator of image-to-CAD reconstruction, 
whose goal is to reconstruct a corresponding CAD model from a reference image. 
Judge the candidates internally and return only the requested top-three ranking line.

\vspace{0.5em}
\textbf{Evaluation Instruction.}
Rank the three best CAD candidates by their overall fidelity to the reference image.

Considering the reference image, assess how effectively each candidate reconstructs 
the depicted object based on the following factors:

\begin{itemize}[leftmargin=1.5em, itemsep=1pt, topsep=3pt]
    \item overall similarity to the reference image and the aesthetic quality of the reconstructed model;
    \item overall silhouette, dominant shape, and global proportions;
    \item the number, shape, relative size, position, orientation, arrangement, and connection of major components;
    \item important geometric relationships, including alignment, spacing, symmetry, attachment, support, and containment;
    \item visible local CAD features such as holes, slots, openings, recesses, and protrusions;
    \item missing major structures, unsupported additions, floating parts, and severe unintended intersections.
\end{itemize}

\textbf{Image Layout.}
\begin{itemize}[leftmargin=1.5em, itemsep=1pt, topsep=3pt]
    \item The top-left image is the reference image. It is used only as the target and must not be ranked.
    \item The five CAD candidates are labeled by position in reading order: 
    \texttt{a} is top-center, \texttt{b} is top-right, 
    \texttt{c} is bottom-left, \texttt{d} is bottom-center, 
    and \texttt{e} is bottom-right.
\end{itemize}

\textbf{Output Rules.}
\begin{itemize}[leftmargin=1.5em, itemsep=1pt, topsep=3pt]
    \item Select and rank only the three best candidates, with no ties.
    \item Do not provide analysis, explanations, headings, markdown, or any other text.
    \item Return exactly one line in the following format:
\end{itemize}

\begin{center}
\texttt{TOP 3: 1=a, 2=b, 3=c}
\end{center}

Replace the letters with the actual ordering. Use three distinct letters from \texttt{a--e}.

\end{tcolorbox}

For the VLM-based evaluation, we arrange the reference image and the rendered outputs of the five evaluated methods into a \(2\times3\) image sheet and present it to Gemini-3.1-Pro-Preview. Following the evaluation criteria, the model assesses the candidates holistically and selects the exact Rank-1, Rank-2, and Rank-3 results that best reconstruct the object depicted in the reference image.

For the human user study, we develop a web-based interface that presents the same reference image and five rendered results in each trial, as shown in Fig.~\ref{fig:webpage}. Method identities are anonymized, and the display order is randomized to reduce potential bias. Participants assess the candidates according to the guidelines and select their exact Rank-1, Rank-2, and Rank-3 choices for each case.

\subsection{Implementation Details}
\label{sup:imple}
All experiments are conducted on a server equipped with two Intel Xeon Platinum 8350C CPUs and an NVIDIA A800 GPU with 80 GB of memory. For general-purpose VLMs, we follow the official documentation and recommended inference settings. Each model is provided with the input image and instructed to generate the corresponding CadQuery code. Since direct VLM generation and CAD-Coder may occasionally produce non-executable programs, we repeatedly query these methods until an executable CAD program is obtained, ensuring that every evaluation sample has a valid output for geometric comparison.  

For all baselines, we use the officially released checkpoints and follow the default configurations provided in their respective repositories. For agentic methods, including Procedura, ArtiCraft, and CADForge, we use GPT-5.5 as the backbone model and retain their native iterative regeneration and repair mechanisms. For CADForge, we allow up to five code-repair attempts and three review-triggered revisions for each component-generation step. 

For implementation convenience, we also equip the Geometry Program Agent with a small deterministic library of reusable utility functions for common geometric computations in CAD. These utilities are simple functional wrappers, introduce no additional learned components, and can be equivalently implemented using local mathematical operations.

After each successful component-level CadQuery execution, CADForge exports the updated model in STEP, STL, and GLB formats and use the GLB representation for visual review. The model is rendered off-screen with VTK from two viewpoints: a front view at an elevation of \(10^\circ\) and an azimuth of \(-90^\circ\), and a front-biased oblique view at an elevation of \(30^\circ\) and an azimuth of \(-70^\circ\). The camera is directed toward the center of the model’s bounding box with a fixed \(z\)-up convention, while its distance is adjusted according to the model extent. We use opaque Phong-shaded surfaces, a white background, explicit feature edges, and three-point illumination without cast shadows. Previously generated components are rendered in gray, whereas the component introduced at the current step is highlighted in blue. Finally, the reference image and the two CAD renderings are resized without cropping or distortion and concatenated into a single review image for direct visual comparison.

For toolkit construction, we use a separate set of 60 cases from the Fusion 360 dataset. Each generated helper is validated for syntactic correctness and retained only if its application improves reconstruction performance on the corresponding failure cases. Through this process, CADForge develop a final toolkit consisting of 10 tools.

For qualitative visualization, we adopt a unified CAD-style rendering protocol across all methods. For each input image, the generated models are rendered using the same camera configuration, white background, lighting setup, and soft-aqua material color with an RGB value of ((189, 214, 210)). For methods such as Articraft that may generate textured or colored meshes, we remove the original textures and materials before rendering. This allows the comparison to focus on geometric structure, including the overall shape, component layout, proportions, holes, cutouts, and local CAD features, rather than differences in appearance or rendering style.

\input{Tex/webpage}

\subsection{Introduction of the Toolkit}
We implement the self-constructed toolkit as a lightweight library of reusable Python helper functions that encapsulate common yet failure-prone CadQuery operations, such as primitive placement, path-based tubular construction and solid composition. The helpers are prepended to the execution namespace of each generated program, allowing the Code Agent to invoke them directly without additional imports. To expose the toolkit to the LLM, we include a compact API manifest in the prompts for initial code generation, component synthesis, and code repair. The manifest provides each helper’s name, function signature, intended functionality, and concise usage examples, together with lightweight guidance for function selection. Since only this compact interface description is exposed to the Code Agent, incorporating new tools introduces only negligible additional token overhead. The Code Agent is encouraged to invoke a helper whenever it matches the required modeling operation, while retaining native CadQuery primitives for geometries not covered by the toolkit. In this way, the toolkit abstracts away fragile low-level implementation details while preserving the flexibility of general CadQuery modeling.

Below, we present a simple helper and its specific implementation. Rounded boxes are a common pattern in CAD modeling and are typically constructed by creating a box and filleting its edges. Directly asking the Code Agent to repeatedly synthesize low-level operation chains such as \texttt{.box(...).edges(...).fillet(...)} can be fragile, as errors may arise from incorrect API composition, inappropriate fillet radii, or kernel-level fillet failures. CADForge therefore introduces a reusable \texttt{rounded\_box} helper that encapsulates this operation sequence, constrains the fillet radius to a geometry-aware safe range, and falls back to the original valid solid when filleting fails. This helper provides a robust primitive for common mildly rounded components such as panels, drawer fronts, cabinet doors, and seat cushions.



\begin{lstlisting}[
    xleftmargin=0.06\linewidth,
    xrightmargin=0.06\linewidth,
    language=Python,
    caption={Implementation of the \texttt{rounded\_box} and \texttt{safe\_fillet} helpers.},
    label={lst:rounded_box},
    basicstyle=\ttfamily\small,
    keywordstyle=\color{blue!70!black},
    stringstyle=\color{green!40!black},
    commentstyle=\color{gray},
    backgroundcolor=\color{gray!5},
    frame=single,
    rulecolor=\color{gray!35},
    framesep=6pt,
    showstringspaces=false,
    columns=fullflexible,
    keepspaces=true
]
def safe_fillet(obj, radius, selector=None):
    try:
        radius = float(radius)
        if radius <= 0:
            return obj
        target = obj.edges(selector) if selector else obj.edges()
        return target.fillet(radius)
    except Exception:
        return obj

def rounded_box(center, size, radius):
    solid = box_at(center, size)
    sx, sy, sz = [abs(float(v)) for v in size]
    limit = max(1e-6, min(sx, sy, sz) * 0.45)
    return safe_fillet(
        solid, min(abs(float(radius)), limit)
    )
\end{lstlisting}

\section{Extension and Application}
\subsection{Texture Generation and Articulation Modeling}
Since CADForge represents an object as part-wise B-rep geometry with executable modeling code, its output can be readily augmented with additional appearance and articulation attributes. By incorporating two LLM-based agents for texture assignment and articulation modeling, this representation can be extended to support textured and articulated 3D asset generation. Specifically, the texture agent assigns PBR materials to individual parts, while the articulation agent identifies movable parts and predicts their joint types, motion axes, and motion ranges. As shown in Fig.~\ref{fig:case_texture_articulation}, CADForge more faithfully preserves fine-grained part structures than ArtiCraft, such as the number and arrangement of shelves, while maintaining proper connectivity between movable parts and their supporting structures. These results further demonstrate the flexibility of CADForge's structured representation in supporting appearance and articulation modeling beyond geometric reconstruction.

\subsection{Editing Flexibility}
The code-based representation of CADForge naturally provides editability as an inherent property of the generated parametric CAD models. While our primary focus is single-view CAD reconstruction rather than CAD editing, the generated programs can be further modified through natural-language instructions, which can be useful for downstream applications such as scene generation (\cite{yang2025mmgdreamer,yang2026flowscene,xia2026sage,wang2026scenecode}) and spatial intelligence (\cite{wang2025spatial, ma2025ctsg}. By leveraging an LLM to interpret both the current CAD state and an editing request, users can perform global or local modifications while preserving the underlying parametric structure. This also enables convenient exploration of design variants derived from an existing reconstruction. As a qualitative demonstration, Fig.~\ref{fig:case_edit} shows a sequence of edits to a reconstructed truss, including doubling its overall length, reversing the direction of the diagonal braces, and adding additional sections. This example qualitatively demonstrates the editability of the generated code-based CAD representation, while our evaluation focuses on the primary task of single-view CAD reconstruction.

\section{Additional Experiments}

\subsection{Per-benchmark Quantitative Analysis} 
We further report the performance on each individual subset. As shown in Tab.~\ref{tab:comparison_three_benchmarks}, CADForge maintains a clear advantage across the three datasets. Interestingly, on Fusion 360, strong VLMs such as GPT-5.6-Sol are highly competitive. This may be attributed to the relatively regular and structurally simple geometries in Fusion 360, for which direct program generation is already effective. In contrast, on more compositionally complex datasets such as 3D-FUTURE, the advantage of CADForge over its GPT-5.5 backbone and even the stronger GPT-5.6-Sol becomes more pronounced. This suggests that explicit geometry reasoning and progressive component-wise construction provide greater benefits as geometric complexity and inter-component dependencies increase.
\input{Tables/results_benchmark}

\subsection{Additional Ablation on Geometry Programming}
To further evaluate the role of Geometry Programming within the Geometry Reasoning process, we conduct an ablation study on 100 randomly selected cases, including 30 from Objaverse, 30 from Fusion 360, and 40 from 3D-FUTURE. For the CADForge variant without Geometry Programming, it directly predicts the key parameters from the geometric specification. As shown in Tab.~\ref{tab:ablation_geometry_programming}, removing Geometry Programming consistently degrades performance across all metrics. These results indicate that explicitly translating geometric specifications into executable programs provides more reliable and precise modeling parameters than direct parameter prediction, leading to improved geometric fidelity and structural consistency.

\subsection{User Study}
\label{user_study}
Following the evaluation protocol in Sec.~\ref{criteria}, we also conduct a human user study with five evaluators. Each evaluator independently assesses all samples in the evaluation set based on the rendered results and assigns exact Rank-1, Rank-2, and Rank-3 choices to the three preferred methods. For each evaluator, we first compute their the exact Rank-$k$ preference rate ($k=1,2,3$) of each method. We then average the resulting preference rates across all the evaluators. As shown in Tab.~\ref{tab:user_study}, CADForge demonstrates a clear advantage under human assessment, consistent with the geometric and VLM-based preference evaluations reported in the main paper.
\input{Tables/ablation_geometry_programming}
\input{Tables/user_study}

\subsection{Additional Qualitative Comparison}
\input{Tex/case_sup}
\textbf{Generation Results from Different Methods}. 
Fig.~\ref{fig:case_sup} presents additional qualitative comparisons across different methods. We also include cadrille (\cite{kolodiazhnyi2025cadrille}), using point clouds sampled from the ground-truth shapes as input. In the first row, CADForge faithfully preserves the curved tubular structure and component connectivity, while Articraft exhibits noticeable geometric distortions and GPT-5.6-Sol leaves the top tubular component disconnected. In the third row, CADForge better captures the thin supporting frames and shelf layout, whereas other methods often introduce misplaced or redundant structures. In the last row, it more accurately recovers the cabinet's complex internal layout and partitioning. Notably, CADForge achieves better reconstruction quality than cadrille despite relying solely on a single input image. 

\input{Tex/case_texture_articulation}
\input{Tex/case_edit}

\textbf{Performance of CADForge across Different Backbones}. To further assess the generality and robustness of our framework, we visualize CAD models produced by CADForge with different backbone models, together with results from two representative agentic methods, Articraft and Procedura. As shown in Fig.~\ref{fig:case_backbones}, when all using GPT-5.5, CADForge achieves higher structural fidelity than both baselines. For the challenging object in the first row, featuring intertwined curved tubes and slender bars, CADForge better preserves the overall topology and component layout, whereas Articraft exhibits noticeable geometric distortions and Procedura produces an overly entangled lower structure. In the second row, CADForge more faithfully recovers the pedestal-table structure and its part proportions, while the competing methods introduce incorrect proportions or less accurate component geometries. Moreover, CADForge performs consistently well across different backbone models, while stronger backbones may lead to further improvements in reconstruction quality.

\textbf{Visualization of Component Decomposing}. As shown in Fig.~\ref{fig:case_decompose}, CADForge produces reasonable and concise decomposition plans for both cases. Each component has clear functional semantics, and the generation order follows practical CAD modeling logic. Guided by the decomposition plan, CADForge accurately captures the geometric structure of each component and generates new parts that are well aligned and geometrically consistent with the existing ones, ultimately producing coherent and structurally sound CAD models. For the two cases shown in the figure, Procedura decomposes them into 33 and 14 parts, respectively, leading to overly fragmented plans. Such fine-grained decomposition increases the number of generation steps and token costs. More importantly, it causes closely related parts to be generated separately across different steps, making their alignment, connection, and assembly relationships more difficult to model. In contrast, CADForge strikes a better balance between structural integrity and decomposition granularity, producing more compact and coherent plans that facilitate subsequent CAD generation.

\section{Limitations and Future Work}
Our current framework primarily focuses on single-view image-conditioned CAD reconstruction. While this setting is practically useful, it would be beneficial to extend CADForge to incorporate multi-view observations, which could provide richer geometric constraints and further improve reconstruction accuracy and consistency.
In addition, current LLM/VLM backbones still have limited 3D perception capability under extreme viewing conditions. Consequently, CADForge may fail to recover complete object geometry from highly unfavorable viewpoints, such as near-complete side or rear views. Future work could explore stronger view-aware reasoning or mechanisms for explicitly modeling geometric uncertainty under severely incomplete observations.

\section{Prompt Templates}

\input{Tex/case_decompose}
\input{Tex/case_backbones}

\subsection{Component Decomposing}
\begin{tcolorbox}[
    title={Prompt for Initial Component Decomposing},
    colback=gray!5,
    colframe=gray!60,
    boxrule=0.6pt,
    arc=2pt,
    left=6pt,
    right=6pt,
    top=6pt,
    bottom=6pt,
    fonttitle=\bfseries,
    breakable
]
\label{prompt:initial_object_planning}

\textbf{System Prompt:}
\vspace{0.5em}

You are the Initial Object Planner for image-to-CAD reconstruction.

\vspace{0.5em}
\textbf{Task.}
Identify the depicted object and produce an ordered inventory of the semantic
components to be constructed. Do not estimate dimensions or write CAD code.

Visible image evidence must be treated as the primary source of information.
Use high-confidence object knowledge only to recover necessary occluded external
counterparts, connectivity, or support. Do not invent unseen internal,
underside, or manufacturing structures.

\vspace{0.5em}
\textbf{Output Format.}
Return only the following fields:

\begin{itemize}[leftmargin=1.5em, itemsep=1pt, topsep=3pt]
    \item \texttt{object\_summary}: one concise but precise description of the
    complete object. Prioritize geometric truth over semantic interpretation.
    \item \texttt{component\_inventory}: an ordered list of
    \texttt{\{component\_id, brief\}} entries. The first entry is simply the
    semantic component to be constructed first.
\end{itemize}

\textbf{Inventory Rules.}
\begin{itemize}[leftmargin=1.5em, itemsep=1pt, topsep=3pt]
    \item Treat each inventory item as a CAD-meaningful component or coherent
    subassembly, rather than an individual primitive or, by default, the entire
    object.

    \item Do not repeat the same physical part or feature across multiple
    entries. Assign it to exactly one inventory item, and ensure that subsequent
    entries introduce only new components.

    \item Every inventory item must be independently addable positive geometry.
    A hole, slot, bore, recess, opening, groove, or any other subtractive feature
    must not be listed as a separate workflow item. Include it in the brief of
    its owning component and construct it together with that host.

    \item Group members whose correctness depends on a shared envelope, profile,
    axis, spacing pattern, or collective silhouette into one coherent component,
    even when they form separate solids.

    \item When a visible frame, lattice, support skeleton, rail network, or other
    continuous structural framework is primarily defined by its overall outline
    and internal pattern, represent it as one coherent framework component when
    this is the natural CAD unit.

    \item Separate physical pieces only when their geometry can be determined
    and reviewed without relying on a shared multi-member layout.

    \item Omit non-structural visual effects and fine decorative lines, including
    lighting strips, glow bands, texture seams, color trim, and printed outlines,
    unless they define a physical cut, handle, rail, panel boundary, or support.

    \item Order the inventory from a stable geometric basis to dependent
    components and details. For the first component, prefer the visible
    structural component that provides the most reliable relevant-axis extents
    and useful interfaces for subsequent placement. It may be volumetric or
    planar and must not be selected merely because it is the largest component.

    \item Assign every item a unique and stable semantic
    \texttt{component\_id}, including the first item. Never use sequence-only
    identifiers such as \texttt{initial\_component} or \texttt{part\_1}.

    \item Each \texttt{brief} must describe, in one or two sentences, the
    component's semantic role, dominant shape or profile, required grouping or
    count, clearly image-supported material-versus-empty topology, and
    relationship to earlier components. Do not prescribe uncertain dimensions,
    internal subdivisions, or boundary directions; the Component Observer
    determines detailed geometry from the image.
\end{itemize}

Cover every visually or functionally essential component exactly once. Omit
material, color, styling prose, numeric geometry, optional decoration, and
extended discussions of uncertainty.

\vspace{0.5em}

\textbf{User Prompt:}
\vspace{0.5em}

Produce the initial ordered object plan for progressive CAD construction.
Do not include per-step geometry.

\end{tcolorbox}

\subsection{Geometry Observation}
\begin{tcolorbox}[
    title={Prompt for Component Geometry Observation},
    colback=gray!5,
    colframe=gray!60,
    boxrule=0.6pt,
    arc=2pt,
    left=6pt,
    right=6pt,
    top=6pt,
    bottom=6pt,
    fonttitle=\bfseries,
    breakable
]
\label{prompt:component_geometry_observation}

\textbf{System Prompt:}

\vspace{0.5em}
You are the Component Geometry Observer.

\vspace{0.5em}
\textbf{Inputs and Scope.}
The inputs consist of the original image, one fixed
\texttt{target\_component} from the ordered inventory, a semantic catalog of
\texttt{existing\_parts}, and optional \texttt{key\_notes} provided during
review correction. Do not select the next component or write a modeling recipe.

Treat visible image evidence as the primary source of information. Use
high-confidence object knowledge only to recover functionally necessary
occluded counterparts, connectivity, support, or containment. Mark such facts
as inferred, and do not invent optional details.

\vspace{0.5em}
\textbf{Output Format.}
Return only the following fields:

\begin{itemize}[leftmargin=1.5em, itemsep=1pt, topsep=3pt]
    \item \texttt{reference\_component\_ids}: zero to two identifiers selected
    from \texttt{existing\_parts}.

    \item \texttt{component\_brief}: a compact but precise semantic and
    topological description of the complete target component. For a complex
    single part, include its distinct construction-critical substructures in
    this whole-part description.

    \item \texttt{geometry\_brief}: compact CAD-relevant geometric facts and
    relations. For each named substructure, separately state its useful size or
    profile, placement or spacing, orientation or endpoints, and connection.
\end{itemize}

\textbf{Observation Priority.}
Observe the complete target component before selecting references. Prioritize
applicable information in the following order:

\begin{enumerate}[leftmargin=1.7em, itemsep=2pt, topsep=3pt]
    \item \textbf{Relations to existing parts and overall placement.}
    First describe visible spatial relations to useful existing parts, followed
    by the target's overall placement and occupied extent. Identify the relevant
    reference boundary, face, center, axis, or extent, and state whether the
    target is aligned, inset, offset, spanning, contacting, supporting,
    enclosing, or separated by visible clearance. When the target lies within a
    bounded reference extent, express its relevant center or endpoint as one
    normalized fraction along that same-axis extent. Place these relations at
    the beginning of \texttt{geometry\_brief}, before local dimensions.

    \item \textbf{Component structure.}
    Describe count or grouping, symmetry, dominant shape, and
    material-versus-empty topology, including holes, openings, shells, or
    recesses.

    \item \textbf{Local construction geometry.}
    Describe the main proportions, profile, cross-section, thickness, curvature,
    tilt, and end form.
\end{enumerate}

\textbf{Geometry Rules.}
\begin{itemize}[leftmargin=1.5em, itemsep=1pt, topsep=3pt]
    \item The global CAD frame: right is
    \(+x\), up is \(+z\), and depth from the visible front into the object
    is \(+y\). Preserve this mapping throughout and never swap axes to follow a
    component's local long axis. Express every local direction as a vector in
    the global CAD frame.

    \item When the object has a clear regular frame, infer its canonical axes
    before measuring. Express geometry in object coordinates rather than raw
    perspective projection, while preserving visible asymmetry when it appears
    intentional.

    \item Use the coordinate frame already stored in the selected references.
    Do not redefine the axes. If semantic axes are useful, map them to the
    existing CAD axes.

    \item Use a same-axis reference extent whenever available: width for
    horizontal placement, depth for front--back placement, and height for
    vertical placement. Use another visible scale only when no relevant
    same-axis reference exists.

    \item When \texttt{existing\_parts} is empty, make the observation
    self-contained in the global CAD frame. Normalize one reliable target
    extent to \(1\), and state the target's cross-axis proportions, boundaries,
    and internal layout. Other visible but unbuilt components may inform
    perception, but they must not become numerical dependencies or receive
    proxy dimensions.

    \item State one best estimate for every useful ratio, count, angle, offset,
    depth, radius, or spacing. Never output numeric ranges.

    \item When the target contributes to a visible outer boundary of the object
    or a subassembly, state whether its extreme edge is aligned with, inset from,
    or extends beyond the corresponding reference edge, using one exact offset
    or ratio. Fix this outer boundary before estimating the internal layout.

    \item For a tilted or curved element, identify its plane together with an
    angle, endpoint offset, radius, rise, or another construction-usable
    descriptor.

    \item When contact, support, coverage, insertion, or containment applies,
    identify the participating faces, edges, or extent boundaries and state
    their intended relation. Do not reduce a boundary relation to vague center
    proximity or invent contact where the image indicates clearance.

    \item Determine open-versus-closed topology independently at each boundary.
    Do not treat an occluded or unseen end as open. Retain a simple closure when
    required by the component's evident containing or supporting function,
    unless visible geometry or semantics indicate a through-opening.

    \item Distinguish directly visible facts from inferred necessary structure.

    \item Image measurements override generic proportions or numerical values
    stated in prior text.

    \item A reference may provide contact, scale, center, axis, alignment,
    envelope, or symmetry. Select it only when its geometry is required as a
    numerical reference downstream.
\end{itemize}

\textbf{Output Organization and Corrections.}
Place shape and topology facts in \texttt{component\_brief}, and place
quantitative or explicit spatial relations in \texttt{geometry\_brief}. Begin
\texttt{geometry\_brief} with the target's overall placement and extent, followed
by its internal layout and local measurements. An answer that only names the
component or omits an observable construction-critical relation is incomplete. When \texttt{key\_notes} are provided, re-examine the image and return a complete
replacement observation incorporating only corrections supported by the image.
Do not merely paraphrase.

\vspace{0.5em}
\textbf{User Prompt (Observe):}
\vspace{0.5em}

Observe the fixed target component and select only completed components whose
stored parameters provide useful references:
\vspace{0.5em}
\ttfamily
\{\\
\quad ``target\_component'': <TARGET\_COMPONENT\_FROM\_INVENTORY>,\\
\quad ``existing\_parts'': <COMPLETED\_EXISTING\_PARTS>\\
\}
\end{tcolorbox}

When the Reviewer Agent determines that an unsatisfactory construction attempt is
caused by the geometry module, the target component is re-observed using the
following user prompt:

\begin{tcolorbox}[
    title={User Prompt for Geometry Re-observation},
    colback=gray!5,
    colframe=gray!60,
    boxrule=0.6pt,
    arc=2pt,
    left=6pt,
    right=6pt,
    top=6pt,
    bottom=6pt,
    fonttitle=\bfseries,
    breakable
]
Re-observe the fixed target component from the original image and return a
complete replacement observation. Select only completed components whose stored
parameters provide useful geometric references:
\vspace{0.5em}
\ttfamily
\{\\
\quad ``target\_component'':
\textless TARGET\_COMPONENT\_FROM\_INVENTORY\textgreater,\\
\quad ``existing\_parts'':
\textless COMPLETED\_EXISTING\_PARTS\textgreater,\\
\quad ``key\_notes'':
\textless REVIEW\_CORRECTION\_NOTES\textgreater\\
\}

\end{tcolorbox}

\subsection{Geometry Programming}

\begin{tcolorbox}[
    title={Prompt for Geometry Program Generation},
    colback=gray!5,
    colframe=gray!60,
    boxrule=0.6pt,
    arc=2pt,
    left=6pt,
    right=6pt,
    top=6pt,
    bottom=6pt,
    fonttitle=\bfseries,
    breakable
]
\label{prompt:geometry_program_agent}

\textbf{System Prompt:}
\vspace{0.5em}

You are the Geometry Program Agent.

\vspace{0.5em}
\textbf{Task.}
Compile the component observation and selected prior-component parameters into a
small, pure-Python geometry solver. The image may be used to resolve a missing
numerical estimate, but do not re-plan the component or generate CadQuery code.

\vspace{0.5em}
\textbf{Required Output.}
Return only \texttt{solver\_code} defining exactly:

\begin{flushleft}
\ttfamily
def solve(reference\_components, tools):\\
\quad ...\\
\quad return \{\ldots\}
\end{flushleft}

\textbf{Coordinate Convention.}
CAD coordinates use \(x\) for object right, \(y\) from the visible front toward
the back, and \(z\) for up. The main visible front therefore normally lies in
the \(XZ\) plane with outward normal \([0,-1,0]\). Respect every explicit frame,
axis, center, boundary, or endpoint specified in the selected component
parameters.

\vspace{0.5em}
\textbf{Solver Rules.}
\begin{itemize}[leftmargin=1.5em, itemsep=1pt, topsep=3pt]
    \item Translate explicit ratios, counts, offsets, spacing, angles, radii,
    symmetry, arrays, centers, axes, endpoints, boundaries, contact relations,
    and coverage relations into constants or formulas.

    \item Convert normalized placement along a reference extent using\begingroup
    \footnotesize
    \[
    \texttt{position}
    =
    \texttt{lower\_boundary}
    +
    \texttt{fraction}
    \times
    (\texttt{upper\_boundary}-\texttt{lower\_boundary}).
    \]
    \endgroup

    \item Resolve overall placement and endpoints from reference boundaries
    before deriving local profile dimensions. Local size ratios must not replace
    an explicit position or extent relation.

    \item Read source values from
    \texttt{reference\_components[id]["parameters"]} when the corresponding
    reference is selected.

    \item Treat only components actually present in
    \texttt{reference\_components} as numerical references. Never invent proxy
    dimensions, bounds, centers, or coordinate frames for an absent or unbuilt
    component.

    \item When \texttt{solve\_mode} is
    \texttt{target\_local\_normalized}, set one reliable extent of the current
    component to \(1\), and solve its remaining dimensions and features from its
    observed proportions in a simple target-local frame. Do not fabricate a
    whole-object envelope to evaluate relations to future components.

    \item Preserve the coordinate frame of the selected references. If an
    observation uses semantic axes, map them to the existing CAD axes before
    performing calculations.

    \item Use one concrete nominal value for each relation and preserve the
    observation's stable reference choice. Prefer same-axis extents only when
    their scale is relevant and comparable. Preserve observed cross-axis
    proportions rather than assigning a default value of \(1\) to several axes.

    \item Realize contact, support, coverage, insertion, and containment through
    boundary or extent equations, such as center plus or minus half-extent,
    rather than through center proximity alone. Solve nominal contact by matching
    the intended local boundaries while preserving intentional gaps and
    clearances.

    \item When a feature has continuous curvature, compute a sufficiently
    sampled centerline or profile that follows the curvature. Do not replace it
    with bounding-box corners or long orthogonal segments unless such straight
    runs are directly observed.

    \item Return nominal geometry only. Do not introduce Boolean tolerances or
    overlaps; the Code Agent is responsible for construction-specific numerical
    tolerances.

    \item Return only values that directly drive the current CAD construction.
    Keep intermediate calculations, copied reference data, derivations,
    tolerances, and duplicate normalized or absolute representations inside
    \texttt{solve()}.

    \item Treat geometry outputs as candidates. The Code Agent determines which
    values are used and which values enter the final
    \texttt{COMPONENT\_PARAMETERS}.

    \item Direct assignments and simple ratios require no tool. Use
    \texttt{tools.\textless name\textgreater(...)} for a matching general
    pattern; otherwise, implement a short local pure-mathematical helper.

    \item When \texttt{repair\_context} is present, repair the supplied solver's
    execution error without changing unrelated geometry.

    \item Do not import modules, invoke CadQuery, access IO, the operating system,
    or the network, or mutate external state.
\end{itemize}

Always return a non-empty dictionary, but keep the solver proportional to the
actual geometric relations: simple geometry should produce simple code.

\vspace{0.5em}
\textbf{User Prompt:}
\vspace{0.5em}

Write the geometry-solving program:

\ttfamily
\begingroup
\footnotesize
\{\\
\quad ``current\_component'':
\textless CURRENT\_COMPONENT\_OBSERVATION\textgreater,\\
\quad ``reference\_components'':
\textless SELECTED\_REFERENCE\_COMPONENT\_PARAMETERS\textgreater,\\
\quad ``solve\_mode'':
``reference\_based'' \textbar{} ``target\_local\_normalized'',\\
\quad ``available\_tool\_api'':
\textless GEOMETRY\_TOOL\_Introduction\textgreater\\
\}
\endgroup
\end{tcolorbox}

\subsection{Code Synthesis}

\begin{tcolorbox}[
    title={Prompt for Component CAD Code Generation},
    colback=gray!5,
    colframe=gray!60,
    boxrule=0.6pt,
    arc=2pt,
    left=6pt,
    right=6pt,
    top=6pt,
    bottom=6pt,
    fonttitle=\bfseries,
    breakable
]
\label{prompt:component_cad_code_agent}

\textbf{System Prompt:}
\vspace{0.55em}

You are the Component CAD Code Agent.

\vspace{0.5em}
\textbf{Task.}
Write one standalone CadQuery script for the requested new component only. Do
not recreate previously accepted components. You do not receive the input image;
\texttt{current\_component} and \texttt{geometry\_outputs} constitute the
image-grounded construction contract.

\vspace{0.5em}
\textbf{Inputs.}
\begin{itemize}[leftmargin=1.5em, itemsep=1pt, topsep=3pt]
    \item \texttt{current\_component}: the semantic and topological description
    together with the observed geometry brief;

    \item \texttt{geometry\_outputs}: solved candidate dimensions and placements;

    \item \texttt{reference\_components}: final public parameters of previously
    completed components, when available;

    \item \texttt{key\_notes}: the latest standalone correction, when available.
\end{itemize}

\textbf{Evidence Priority.}
Prioritize evidence in the following order:
\texttt{key\_notes}, explicit \texttt{geometry\_outputs},
\texttt{current\_component}, reference-component values, and finally
conservative CAD priors for missing details. Use
\texttt{reference\_components} for scale, alignment, placement, and contact
relations, but do not copy their geometry.

\vspace{0.5em}
\textbf{Coordinate Convention.}
CAD coordinates use \(x\) for object right, \(y\) from the visible front toward
the back, and \(z\) for up. The main visible front therefore normally lies in
the \(XZ\) plane with outward normal \([0,-1,0]\). Respect every explicit frame,
axis, center, boundary, or endpoint specified in the selected component
parameters.

\vspace{0.5em}
\textbf{Required Output.}
Return raw Python code only. The script must:

\begin{itemize}[leftmargin=1.5em, itemsep=1pt, topsep=3pt]
    \item import only \texttt{cadquery as cq} and, optionally, \texttt{math};

    \item define \texttt{component\_result} as the completed
    \texttt{Workplane} or \texttt{Shape} for the current step;

    \item define \texttt{COMPONENT\_PARAMETERS} as a compact,
    JSON-serializable dictionary;

    \item guard optional operations and make one clear top-level assignment to
    \texttt{component\_result}.
\end{itemize}

\textbf{Construction Standards.}
\begin{itemize}[leftmargin=1.5em, itemsep=1pt, topsep=3pt]
    \item Construct the complete semantic component with the specified
    silhouette, profile, proportions, depth, grouping or count, and
    material-versus-empty topology.

    \item Preserve all explicit dimensions and spatial relations. Complete only
    genuinely missing details using simple and plausible CAD construction.

    \item Prefer the provided CAD helper functions for common boxes, cylinders,
    curved tube-like rods, handles, rails, rounded boxes, unions, cuts, and
    optional fillets. Use tube helpers only when the component geometry implies
    a path with a round cross-section. Use raw CadQuery when no helper is
    applicable.

    \item The helper functions are defined before the generated code is
    executed. Call them directly by name and do not import them. If
    \texttt{geometry\_outputs} contains a helper-ready block whose key matches an
    available helper, use that helper as the primary construction path. Prefer a
    valid, simpler solid over fragile geometric detail.

    \item Construct the host solid first, and apply all required holes, slots,
    windows, recesses, and openings before returning the completed component.
    Use strict cut behavior for required material removal. Never return cutters
    or void proxies as \texttt{component\_result}; reserve soft cut fallbacks
    for optional shallow details.

    \item Distinguish true through-cuts from one-sided cavities, recesses, and
    storage bays. Unless the construction contract explicitly specifies a
    through-opening, preserve a back wall, end wall, or closed continuation
    behind the visible access opening.

    \item Respect all intended alignment, coverage, and interface relations.
    Components described as mounted, resting, inserted, or attached must meet
    the corresponding host boundary and must not float.

    \item Construct a valid solid before applying rounded corners or softened
    edges. Apply fillets only to selected solid edges, using a radius smaller
    than the local thickness and edge length. Guard optional fillets and
    chamfers with a safe fallback so that failure of a decorative operation
    returns the valid unfilleted solid rather than invalidating the component.

    \item The Code Agent is solely responsible for Boolean tolerances. Only when
    required by an operation, use one construction-specific overlap no larger
    than \(0.001\) times the smaller local interface thickness. Do not use this
    tolerance to bridge placement errors or intentional gaps, and do not record
    it in \texttt{COMPONENT\_PARAMETERS}.

    \item Avoid unsupported features, minor surface-only decoration, visible or
    unintended overlap, detached helper solids, and geometry belonging to later
    inventory items.

    \item Construct \texttt{COMPONENT\_PARAMETERS} from the geometry actually
    created by the code. Retain geometry outputs that are used, add newly
    introduced construction-driving dimensions or placements that may be
    referenced by later components, and omit unused inputs, helper values,
    tolerances, and duplicate representations. Reuse the same Python variables
    in both the construction code and \texttt{COMPONENT\_PARAMETERS}.
\end{itemize}

\textbf{Execution Restrictions.}
Do not export files, access IO, the operating system, or the network, use
external assets, or define global workflow state. The local runner is
responsible for assembling and exporting the final model.

\end{tcolorbox}

The following user prompt is used for constructing a regular component.

\begin{tcolorbox}[
    title={User Prompt for Component CAD Code Generation},
    colback=gray!5,
    colframe=gray!60,
    boxrule=0.6pt,
    arc=2pt,
    left=6pt,
    right=6pt,
    top=6pt,
    bottom=6pt,
    fonttitle=\bfseries,
    breakable
]
\label{prompt:component_cad_code_user}

\begin{flushleft}
\ttfamily
AVAILABLE CAD HELPERS:\\
\textless CAD\_HELPER\_MANIFEST\textgreater
\end{flushleft}

Write the new component CadQuery script:

\begin{flushleft}
\ttfamily
\{\\
\quad ``current\_component'':\\
\qquad \textless CURRENT\_COMPONENT\_OBSERVATION\textgreater,\\
\quad ``reference\_components'':\\
\qquad \textless SELECTED\_REFERENCE\_COMPONENT\_PARAMETERS\textgreater,\\
\quad ``geometry\_outputs'':\\
\qquad \textless GEOMETRY\_PROGRAM\_OUTPUTS\textgreater\\
\}
\end{flushleft}

\end{tcolorbox}

When the Review Agent determines that an unsatisfactory construction attempt is caused by the generated CadQuery code, the code generator is invoked again with the
following user prompt, which additionally includes the review correction notes.

\begin{tcolorbox}[
    title={User Prompt for Corrected Component CAD Code Generation},
    colback=gray!5,
    colframe=gray!60,
    boxrule=0.6pt,
    arc=2pt,
    left=6pt,
    right=6pt,
    top=6pt,
    bottom=6pt,
    fonttitle=\bfseries,
    breakable
]
\label{prompt:component_cad_code_correction}

\begin{flushleft}
\ttfamily
AVAILABLE CAD HELPERS:\\
\textless CAD\_HELPER\_MANIFEST\textgreater
\end{flushleft}

Write the new component CadQuery script:

\begin{flushleft}
\ttfamily
\{\\
\quad ``current\_component'':\\
\qquad \textless CURRENT\_COMPONENT\_OBSERVATION\textgreater,\\
\quad ``reference\_components'':\\
\qquad \textless SELECTED\_REFERENCE\_COMPONENT\_PARAMETERS\textgreater,\\
\quad ``geometry\_outputs'':\\
\qquad \textless GEOMETRY\_PROGRAM\_OUTPUTS\textgreater,\\
\quad ``key\_notes'':\\
\qquad \textless REVIEW\_CORRECTION\_NOTES\textgreater\\
\}
\end{flushleft}

\end{tcolorbox}

\subsection{Code Repair}
\begin{tcolorbox}[
    title={Prompt for CAD Script Repair},
    colback=gray!5,
    colframe=gray!60,
    boxrule=0.6pt,
    arc=2pt,
    left=6pt,
    right=6pt,
    top=6pt,
    bottom=6pt,
    fonttitle=\bfseries,
    breakable
]
\label{prompt:cad_script_repair}

\textbf{System Prompt:}
\vspace{0.5em}

You repair a CadQuery script that has failed static validation or execution.

Return only the complete corrected Python script. Make the smallest technical
repair required, while preserving the original geometry, topology, numerical
intent, and output contract. A first-component script must return
\texttt{result} and \texttt{COMPONENT\_PARAMETERS}; a subsequent-component
script must return \texttt{component\_result} and
\texttt{COMPONENT\_PARAMETERS}.

Do not redesign the object, introduce new features, export files, or access IO,
the operating system, or the network. Prefer the provided CAD helper functions
for fragile operations when applicable.

If a fillet, chamfer, selector, shell, Boolean operation, or optional detail
fails, replace only that operation with a more robust construction or a guarded
fallback. The main solid must still be returned when decorative rounding or
optional detail is skipped.

Never replace a required cut with an additive ring, liner, cap, or cutter solid.

\vspace{0.5em}
\textbf{User Prompt:}
\vspace{0.5em}

Repair the script so that it executes successfully. Prefer robust CadQuery
operations over decorative detail. If the error originates from a fillet,
chamfer, selector, Boolean operation, or other optional operation, guard or
remove only that fragile operation while preserving the main component geometry.

\begin{flushleft}
\ttfamily
AVAILABLE CAD HELPERS:\\
\textless CAD\_HELPER\_MANIFEST\textgreater
\end{flushleft}

\begin{flushleft}
\ttfamily
ERROR:\\
\textless EXECUTION\_OR\_STATIC\_VALIDATION\_ERROR\textgreater
\end{flushleft}

\begin{flushleft}
\ttfamily
PREVIOUS SCRIPT:\\
\textless PREVIOUS\_FAILED\_SCRIPT\textgreater
\end{flushleft}

\end{tcolorbox}

\subsection{Review Agent}

\begin{tcolorbox}[
    title={Prompt for Progressive CAD Step Review},
    colback=gray!5,
    colframe=gray!60,
    boxrule=0.6pt,
    arc=2pt,
    left=6pt,
    right=6pt,
    top=6pt,
    bottom=6pt,
    fonttitle=\bfseries,
    breakable
]
\label{prompt:step_review_agent}

\textbf{System Prompt:}
\vspace{0.5em}

You are the Step Review Agent.

\vspace{0.5em}
\textbf{Review Input.}
Inspect one labeled comparison image arranged from left to right as
\texttt{REFERENCE}, \texttt{CAD FRONT}, and \texttt{CAD OBLIQUE}. The newly
generated component may be highlighted, but its color is only a progress cue.

Evaluate \texttt{current\_component} through direct visual comparison with the
reference image, using previously constructed geometry only as spatial context.
The \texttt{current\_component} field identifies the semantic target; it does not
provide numerical evidence for accepting the CAD result.

Treat visible image evidence as the primary source of information. Use
high-confidence object knowledge only to recover functionally necessary
occluded counterparts, connectivity, support, or containment. Mark such facts
as inferred, and do not invent optional details.

\vspace{0.5em}
\textbf{Rejection Criteria.}
Reject the result only for a clear and material failure in at least one of the
following categories:

\begin{enumerate}[leftmargin=1.7em, itemsep=2pt, topsep=3pt]
    \item \textbf{Geometric inconsistency.}
    The result is visibly inconsistent with the reference image in its dominant
    normalized extent, center or boundary placement, proportions, main
    silhouette or profile, orientation, component-defining count or pattern, or
    material-versus-empty topology.

    When the target belongs to a shared structure, evaluate its completed
    overall extent and coverage rather than only its most recently added member.
    When the target is one coherent part, give additional weight to its overall
    shape, key proportions, internal feature positions, and relative layout.

    \item \textbf{Structural or modeling defect.}
    The CAD contains an unmistakable structural defect, including unsupported
    floating geometry where attachment is clearly intended, gross unintended
    interpenetration or merging that destroys the visible form,
    fragmented or invalid geometry, or reversal of a clearly required opening,
    recess, or through-cut.

    A required void must remove material from the host boundary. An added ring,
    cap, or visible cutter is not an acceptable substitute.
\end{enumerate}

\textbf{Review Procedure.}
\begin{itemize}[leftmargin=1.5em, itemsep=1pt, topsep=3pt]
    \item Before deciding, silently compare the target's dominant normalized
    extent, position, and silhouette between \texttt{REFERENCE} and the CAD
    views using a stable visible feature shared by both. Compare geometric
    ratios rather than canvas occupancy.

    \item For connected, framed, or branched structures, also compare their
    corresponding spans, endpoints, and directions.

    \item When stable visible geometry provides scale, compare the target's key
    extent and endpoint relationships against it. A clear mismatch in length,
    protrusion, gap, or alignment is material rather than a minor proportional
    difference.

    \item For protruding, inset, wall-mounted, suspended, or face-attached
    components, treat a clear depth-placement error, front--back reversal,
    incorrect attachment plane, or opposite protrusion direction as material,
    even when the front silhouette appears plausible.

    \item CAD panels are automatically fitted to the geometry constructed so
    far. If the result contains only a standalone normalized primitive and no
    shared scale feature is available, its canvas occupancy does not determine
    its eventual size in the complete object. Do not reject it for this reason
    alone. Continue to assess its intrinsic proportions, profile, and topology,
    and enforce relative scale once other geometry provides a reference.

    \item Evaluate the largest and most visually evident discrepancy first.
    Account for perspective, occlusion, and rendering limitations.
\end{itemize}

\textbf{Acceptance Tolerance.}
Pass a faithful and usable approximation when it preserves the dominant extent,
placement, silhouette, and material-versus-empty topology. Tolerate minor
differences in proportion, thickness, curvature, styling, small details, and
hidden interfaces when they preserve the component's identity and principal
geometry.

A small seam or shallow local overlap at an otherwise plausible interface is
not sufficient for rejection. Do not infer a contact failure from an ambiguous
view, and do not reject a result merely because the CAD representation is less
detailed than the reference image.

\vspace{0.5em}
\textbf{Verdict Definitions.}
Return \texttt{verdict} as exactly one of the following:

\begin{itemize}[leftmargin=1.5em, itemsep=1pt, topsep=3pt]
    \item \texttt{pass}: no material correction is required;

    \item \texttt{geometry\_error}: the observed or solved size, position, count,
    pattern, frame, direction, or proportion must change. Use this verdict when
    the visible correction requires moving, resizing, or rearranging the
    component;

    \item \texttt{modeling\_error}: the geometric intent is adequate, but its
    CadQuery construction, topology, Boolean operation, or local interface
    realization must change;

    \item \texttt{mixed\_error}: both the geometric intent and CAD construction
    are materially incorrect.
\end{itemize}

\textbf{Output Rules.}
For \texttt{pass}, omit \texttt{key\_notes}. For any other verdict, provide one
to three concise natural-language sentences that state only the forward
constraints the next generation must satisfy. Do not refer to previous code,
previous decisions, or parameter names unless the name is part of
\texttt{current\_component}.

\vspace{0.5em}
\textbf{User Prompt:}
\vspace{0.5em}

Review the latest progressive CAD step:

\ttfamily
\{\\
\quad ``current\_component'':\\
\qquad \textless CURRENT\_COMPONENT\_FOR\_REVIEW\textgreater\\
\}

\end{tcolorbox}

%% file: Tex/dataset_presentation.tex
\begin{figure}[t]
    \centering
    \includegraphics[width=\linewidth]{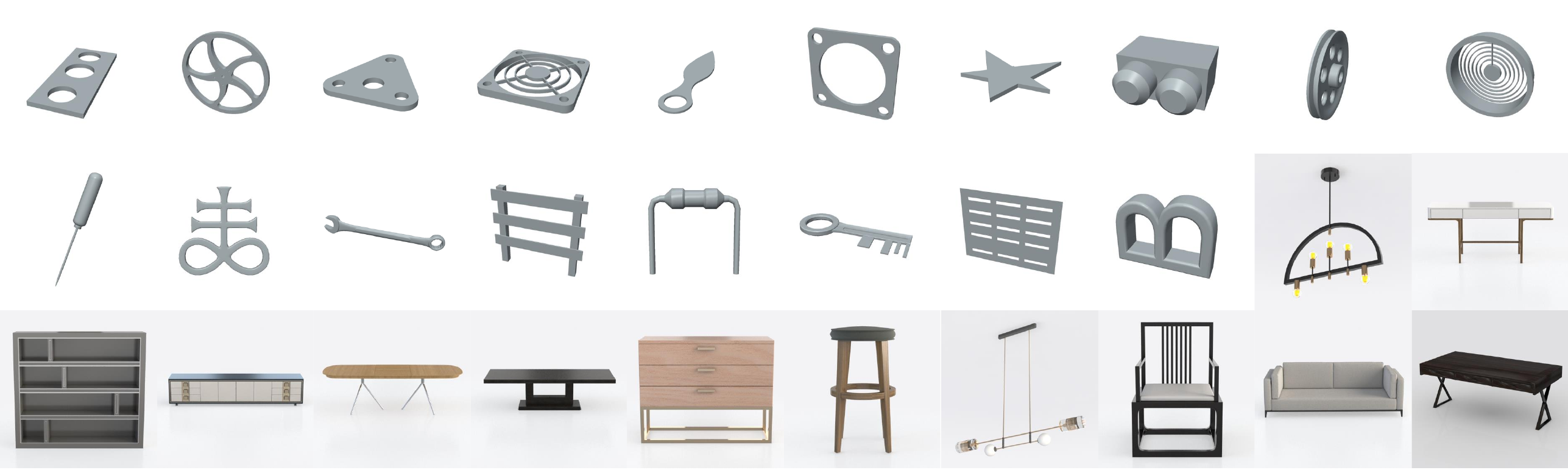}
    \caption{\textbf{Representative input images from our evaluation set}.
    }
    \label{fig:dataset_presentation}
\end{figure}

%% file: Tex/webpage.tex
\begin{figure}[t]
    \centering
    \includegraphics[width=\linewidth]{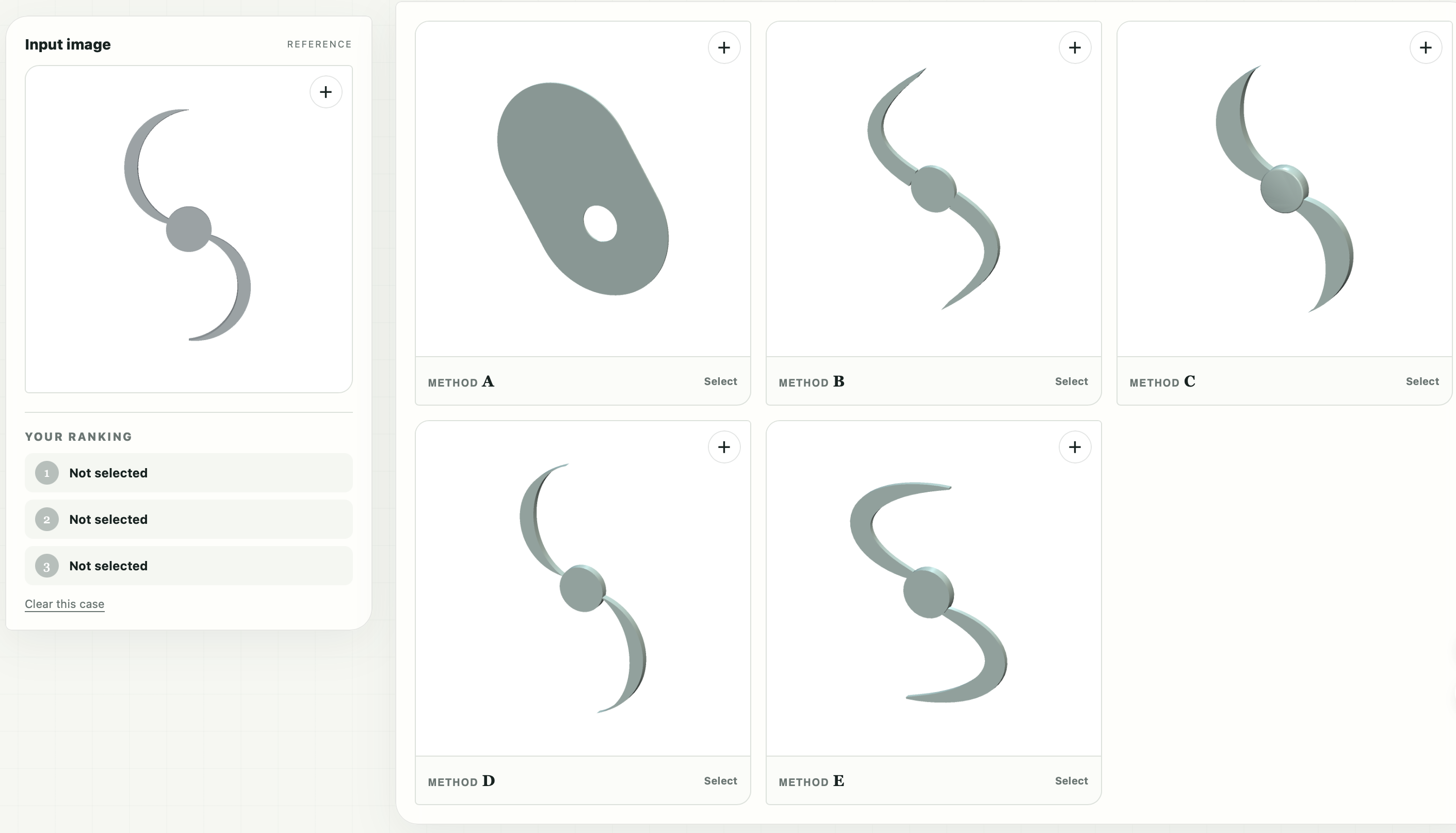}
    \caption{\textbf{Interface for the user study.} In each trial, participants compare the input image with five anonymized and randomly ordered results, and select their Rank-1, Rank-2, and Rank-3 choices.}
    \label{fig:webpage}
    \vspace{-10pt}
\end{figure}

%% file: Tables/results_benchmark.tex
\begin{table}[t]
\centering
\caption{Quantitative comparison across three subsets.
We report CD-L1, Hausdorff Distance (HD), F-scores at thresholds 0.05 and 0.01, Silhouette Intersection over Union (SIOU), and Normal Consistency (NC). All methods are evaluated within each benchmark. }
\label{tab:comparison_three_benchmarks}
\begin{center}
\resizebox{\linewidth}{!}{%
\begin{tabular}{lcccccc}
\toprule
Method
& CD-L1$\downarrow$
& HD$\downarrow$
& F@0.05$\uparrow$
& F@0.01$\uparrow$
& SIOU$\uparrow$
& NC$\uparrow$ \\
\midrule
\multicolumn{7}{l}{\textit{Objaverse Subset}} \\
\midrule
MiniMax-M3
& 0.03155 & 0.17056 & 0.80439 & 0.37348 & 0.59013 & 0.59699 \\
Kimi-K3
& 0.02263 & 0.11200 & 0.89512 & 0.44330 & 0.68132 & 0.67524 \\
GPT-5.5
& 0.02055 & 0.11068 & 0.90861 & 0.44655 & 0.69501 & 0.67405 \\
GPT-5.6-Sol
& 0.01721 & 0.09845 & 0.93483 & 0.49186 & 0.72224 & 0.70549 \\
\cmidrule(lr){1-7}
CAD-Coder
& 0.05017 & 0.21100 & 0.62628 & 0.22364 & 0.50844 & 0.54775 \\
Articraft
& 0.01982 & 0.11151 & 0.92165 & 0.44306 & 0.69408 & 0.66748 \\
Procedura
& 0.01727 & 0.10476 & 0.94058 & 0.47564 & 0.71461 & 0.69639 \\
CADForge (Ours)
& 0.01586
& 0.09089
& 0.94728
& 0.57053
& 0.75019
& 0.71899 \\
\midrule
\multicolumn{7}{l}{\textit{Fusion 360 Subset}} \\
\midrule
MiniMax-M3
& 0.02310 & 0.11499 & 0.87384 & 0.44918 & 0.70372 & 0.63514 \\
Kimi-K3
& 0.01754 & 0.08524 & 0.93066 & 0.41697 & 0.72218 & 0.64841\\
GPT-5.5
& 0.01618 & 0.08206 & 0.94101 & 0.50018 & 0.73900 & 0.64348 \\
GPT-5.6-Sol
& 0.01552
& 0.07342
& 0.94310
& 0.53196
& 0.75389
& 0.66315 \\
\cmidrule(lr){1-7}
CAD-Coder
& 0.02898 & 0.14733 & 0.80716 & 0.37871 & 0.70378 & 0.54979 \\
Articraft
& 0.01641 & 0.08914 & 0.93602 & 0.52808 & 0.75175 & 0.63935 \\
Procedura
& 0.01613 & 0.08110 & 0.94210 & 0.51124 & 0.74224 & 0.65586 \\
CADForge (Ours)
& 0.01541
& 0.07462
& 0.94870
& 0.49131
& 0.75167
& 0.66778 \\
\midrule
\multicolumn{7}{l}{\textit{3D-FUTURE Subset} } \\
\midrule
MiniMax-M3
& 0.02148 & 0.11944 & 0.89588 & 0.43745 & 0.67708 & 0.73874 \\
Kimi-K3
& 0.01909 & 0.10113 & 0.92948 & 0.45340 & 0.72257 & 0.74370 \\
GPT-5.5
& 0.01906 & 0.11184 & 0.91609 & 0.46558 & 0.70969 & 0.72740 \\
GPT-5.6-Sol
& 0.01854 & 0.10450 & 0.92476 & 0.46685 & 0.72758 & 0.75399 \\
\cmidrule(lr){1-7}
CAD-Coder
& 0.05170 & 0.24687 & 0.60815 & 0.22347 & 0.49622 & 0.60783 \\
Articraft
& 0.01697 & 0.10938 & 0.93386 & 0.50799 & 0.74487 & 0.73243 \\
Procedura
& 0.01790 & 0.10939 & 0.93830 & 0.43640 & 0.73569 & 0.72693 \\
CADForge (Ours)
& 0.01601
& 0.09332
& 0.94393
& 0.50962
& 0.75550
& 0.75981 \\
\bottomrule
\end{tabular}%
}
\end{center}
\vspace{-16pt}
\end{table}

%% file: Tables/ablation_geometry_programming.tex
\begin{table}[t]
\centering
\caption{\textbf{Ablation study of the geometry programming}.}
\vspace{-6pt}
\begin{center}
\label{tab:ablation_geometry_programming}
\resizebox{\linewidth}{!}{%
\begin{tabular}{lcccccc}
\toprule
Method
& CD-L1$\downarrow$
& HD$\downarrow$
& F@0.05$\uparrow$
& F@0.01$\uparrow$
& SIOU$\uparrow$
& NC$\uparrow$ \\
\midrule
w/o Geometry Programming
& 0.01655
& 0.09382
& 0.93530
& 0.52369
& 0.74369
& 0.72024 \\
\midrule
\textbf{CADForge}
& \textbf{0.01475}
& \textbf{0.08224}
& \textbf{0.95737}
& \textbf{0.52515}
& \textbf{0.76766}
& \textbf{0.74241} \\
\bottomrule
\end{tabular}%
}
\end{center}
\vspace{-16pt}
\end{table}

%% file: Tables/user_study.tex
\begin{table}[t]
\caption{\textbf{Results of user study.}}
\begin{center}
\label{tab:user_study}
\begin{tabular}{lccc}
\toprule
Method
& Rank-1 (\%)
& Rank-2 (\%)
& Rank-3 (\%) \\
\midrule
CADForge   & 43.80 & 23.80 & 21.10 \\
Procedura & 18.90 & 29.10 & 24.90 \\
GPT5      & 19.90 & 27.20 & 21.50 \\
Articraft & 17.40 & 19.90 & 32.50 \\
CAD-Coder & 0.00  & 0.00  & 0.00 \\
\bottomrule
\end{tabular}
\end{center}
\end{table}

%% file: Tex/case_sup.tex
\begin{figure}[t]
    \centering
    \includegraphics[width=\linewidth]{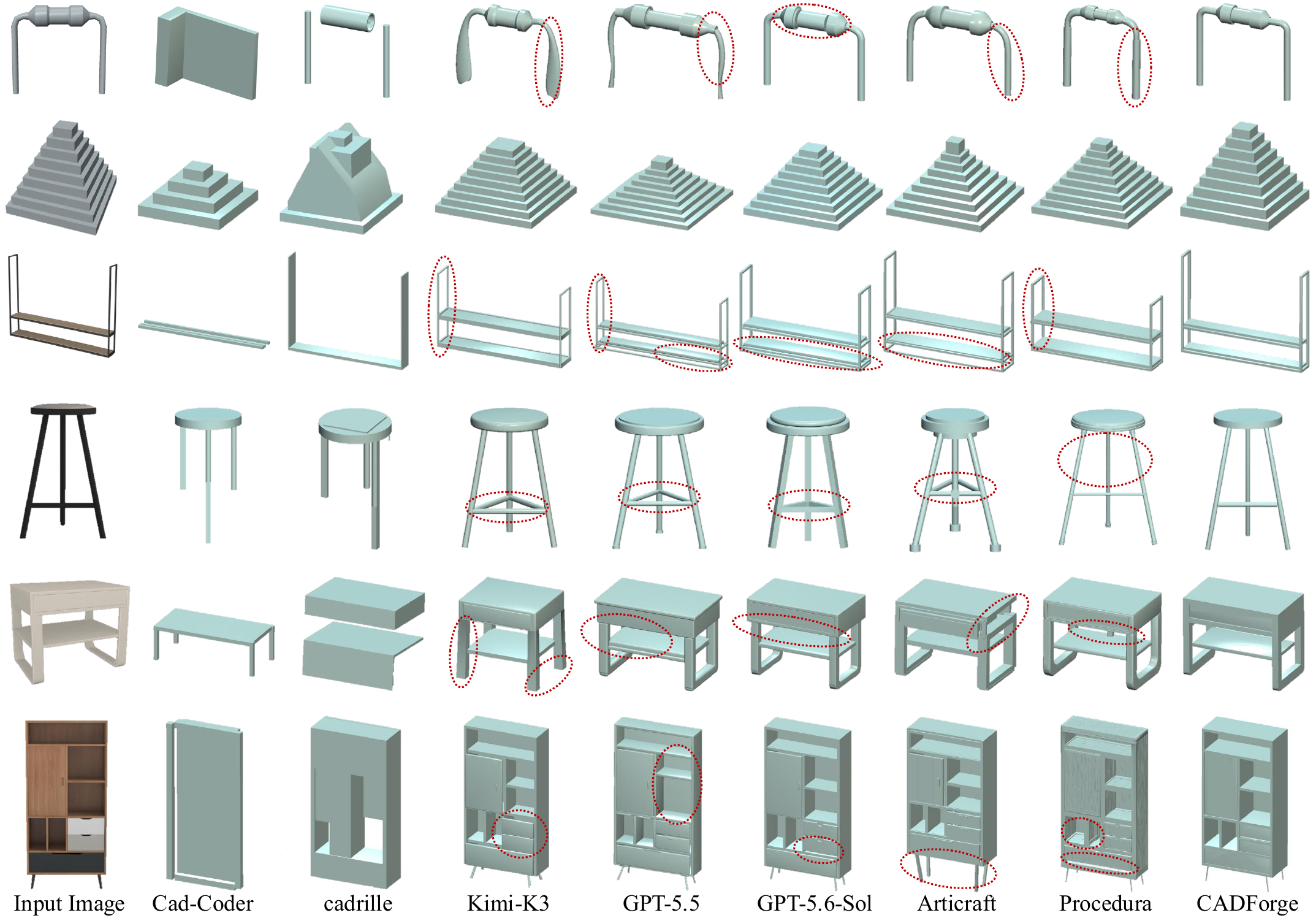}
    \caption{\textbf{Additional qualitative comparison of results generated by different methods}. Red circles highlight regions with geometric inconsistencies relative to the input images.}
    \label{fig:case_sup}
\end{figure}

%% file: Tex/case_texture_articulation.tex
\begin{figure}[t]
\centering
\includegraphics[width=\linewidth]{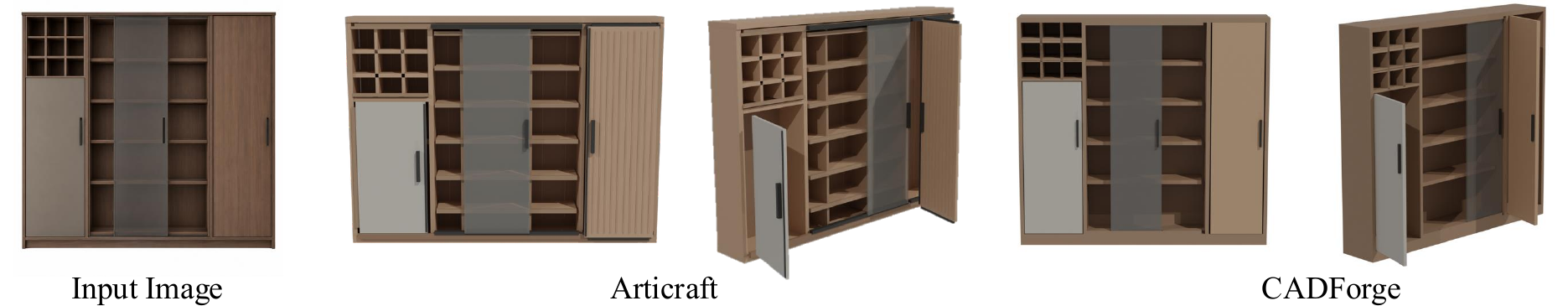}
\caption{\textbf{Extension of CADForge for textured and articulated 3D asset generation}. With additional texture prediction and articulation modeling agents, CADForge can be extended beyond static geometry reconstruction to generate textured and movable 3D assets, while exhibiting improved geometric fidelity and articulation accuracy compared with ArtiCraft. For each asset, the left column shows the rest state (zero articulation), while the right shows the maximum articulation state.
}
\label{fig:case_texture_articulation}
\end{figure}

%% file: Tex/case_edit.tex
\begin{figure}[t]
    \centering
    \includegraphics[width=\linewidth]{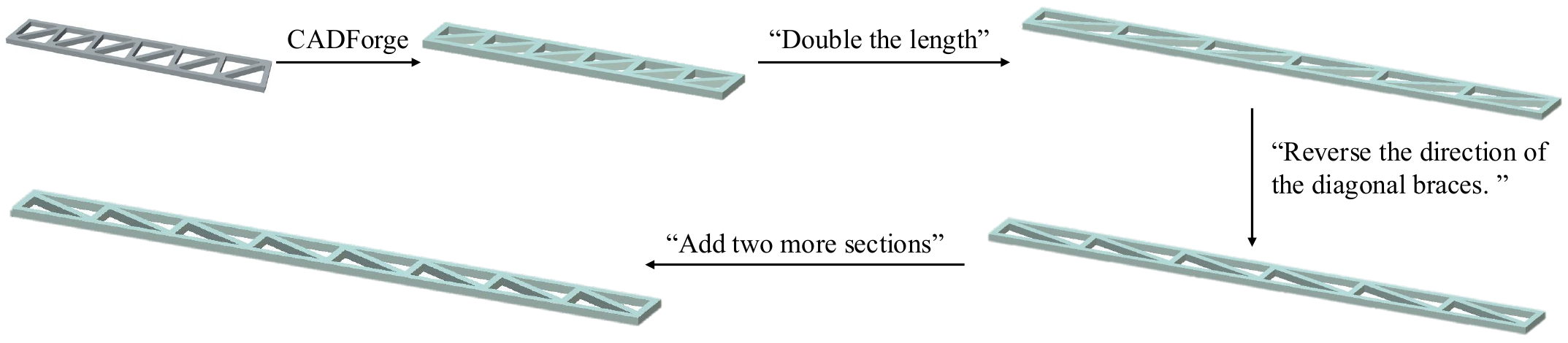}
    \caption{\textbf{Instruction-driven CAD editing with CADForge.} Since CADForge generates part-level CadQuery code, it can be integrated with LLMs to enable natural-language-driven CAD editing. 
    }
    \label{fig:case_edit}
\end{figure}

%% file: Tex/case_decompose.tex
\begin{figure}[t]
    \centering
    \includegraphics[width=\linewidth]{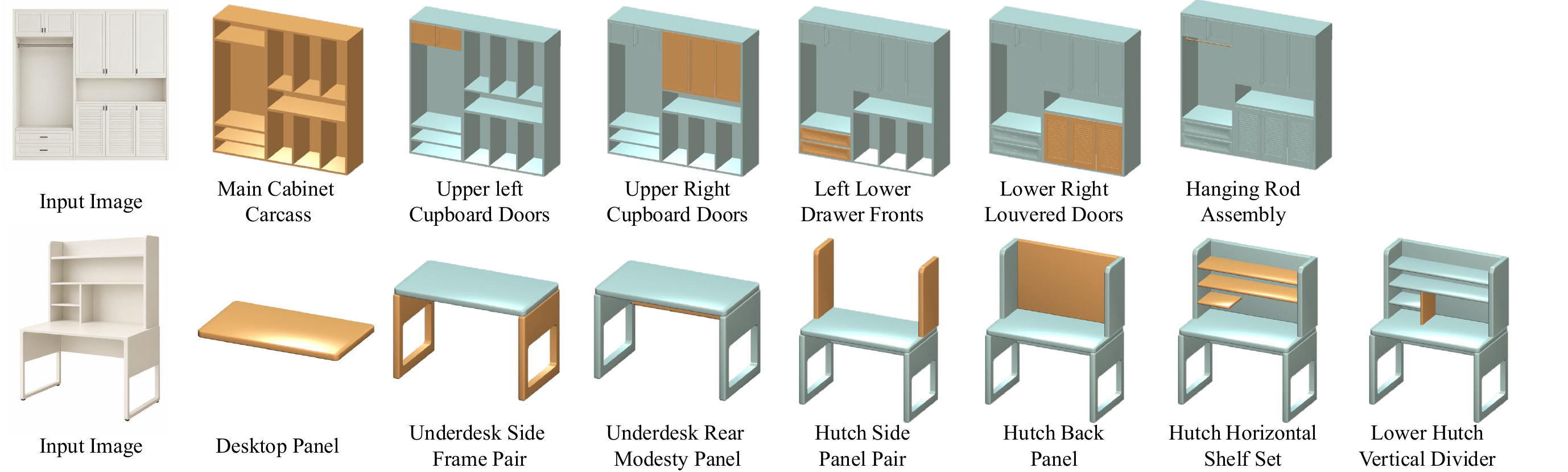}
    \caption{\textbf{Visualization of CAD generation with CADForge on web images.} CADForge decomposes complex CAD generation into a coordinated part-level process, sequentially generating individual parts while maintaining their geometric consistency and relationships to assemble the final model. Newly generated parts are highlighted in yellow, with their part labels shown below.}
    \label{fig:case_decompose}
\end{figure}

%% file: Tex/case_backbones.tex
\begin{figure}[t]
    \centering
    \includegraphics[width=\linewidth]{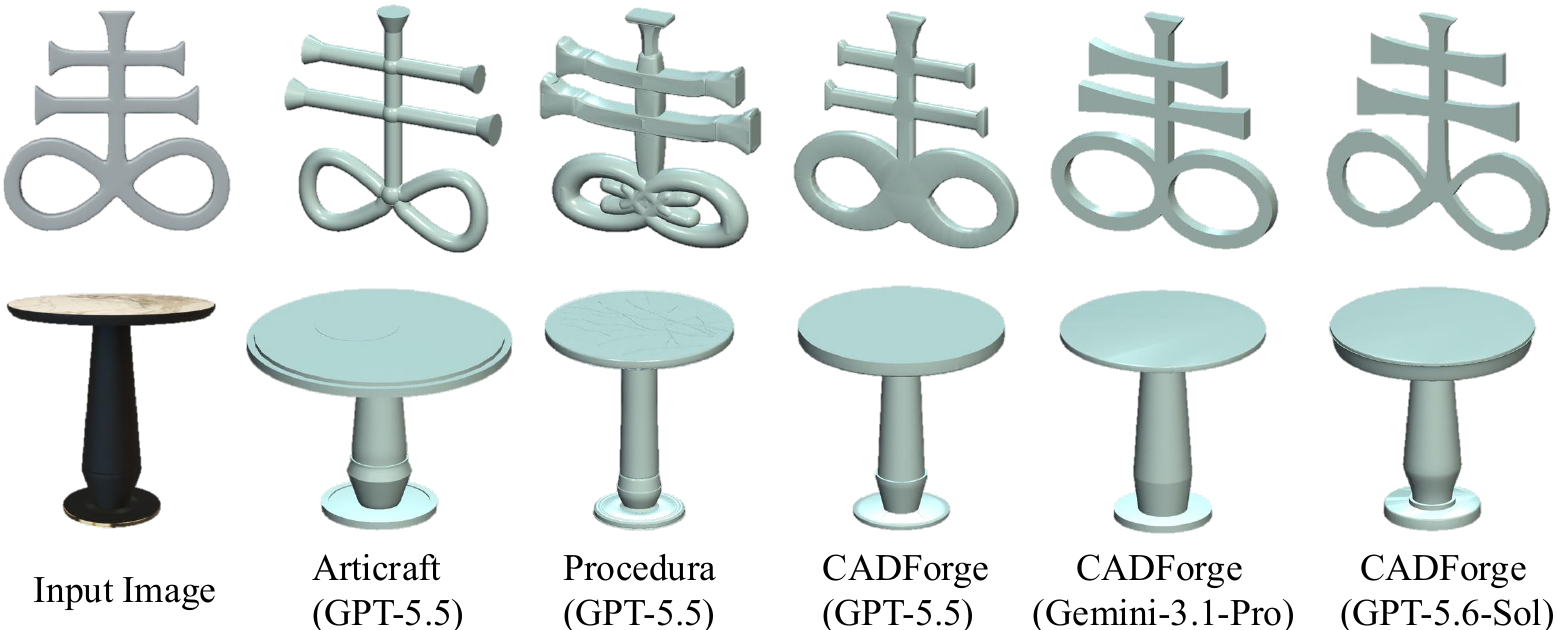}
    \caption{\textbf{Qualitative comparison across different backbone models.} We compare ArtiCraft and Procedura with CADForge instantiated using GPT-5.5, Gemini-3.1-Pro, and GPT-5.6-Sol. CADForge produces structurally consistent reconstructions across different backbones, while stronger backbone models further improve geometric fidelity and shape accuracy.}
    \label{fig:case_backbones}
\end{figure}